\documentclass{article}

    \PassOptionsToPackage{numbers, compress}{natbib}
\usepackage[preprint]{neurips_data_2024}
\usepackage{comment}
\usepackage{paralist}

\usepackage[utf8]{inputenc} % allow utf-8 input
\usepackage[T1]{fontenc}    % use 8-bit T1 fonts
\usepackage{hyperref}       % hyperlinks
\usepackage{url}            % simple URL typesetting
\usepackage{booktabs}       % professional-quality tables
\usepackage{amsfonts}       % blackboard math symbols
\usepackage{nicefrac}       % compact symbols for 1/2, etc.
\usepackage{microtype}      % microtypography
\usepackage{xcolor}         % colors
\usepackage{graphicx}       % nicer figure handling
\usepackage{wrapfig}
\usepackage[inline]{enumitem}
\usepackage{amssymb,multirow}
\usepackage{natbib}
\usepackage{booktabs}
\usepackage{subcaption}
\usepackage{color,tabularx, colortbl,amssymb,amsmath,bm,arydshln}
\usepackage{wrapfig}
\usepackage{array}
\usepackage{float}
\usepackage{listings}
\usepackage{makecell}
\usepackage{amsmath}

\usepackage{tikz}
\newcommand{\tikzxmark}{%
\tikz[scale=0.23] {
    \draw[line width=0.7,line cap=round, color = red] (0,0) to [bend left=6] (1,1);
    \draw[line width=0.7,line cap=round, color = red] (0.2,0.95) to [bend right=3] (0.8,0.05);
}}
\newcommand{\tikzcmark}{%
\tikz[scale=0.23] {
    \draw[line width=0.7,line cap=round, color = green] (0.25,0) to [bend left=10] (1,1);
    \draw[line width=0.8,line cap=round, color = green] (0,0.35) to [bend right=1] (0.23,0);
}}

\usepackage[ruled,vlined]{algorithm2e}

\title{IDAT: A Multi-Modal Dataset and Toolkit for Building and Evaluating Interactive Task-Solving Agents}

\author{%
Shrestha Mohanty$^1$, 
Negar Arabzadeh$^2$, 
Andrea Tupini$^3$,
Yuxuan Sun$^4$,\\
\textbf{
Alexey Skrynnik$^5$,
Artem Zholus$^6$,
Marc-Alexandre Côté$^3$,
Julia Kiseleva$^3$} \\
Massachusetts Institute of Technology$^1$, 
University of Waterloo$^2$,\\
Meta AI$^4$,
AIRI$^5$,
École Polytechnique de Montréal$^6$,
Microsoft Research$^3$\\
shresmoh@mit.edu
}

\begin{document}
\captionsetup{justification=centering} % Center captions
\maketitle

\begin{abstract}
Seamless interaction between AI agents and humans using natural language remains a key goal in AI research. This paper addresses the challenges of developing interactive agents capable of understanding and executing grounded natural language instructions through the IGLU competition at NeurIPS. Despite advancements, challenges such as a scarcity of appropriate datasets and the need for effective evaluation platforms persist. We introduce a scalable data collection tool for gathering interactive grounded language instructions within a Minecraft-like environment, resulting in a Multi-Modal dataset with around 9,000 utterances and over 1,000 clarification questions. Additionally, we present a Human-in-the-Loop interactive evaluation platform for qualitative analysis and comparison of agent performance through multi-turn communication with human annotators. We offer to the community these assets referred to as IDAT (IGLU Dataset And Toolkit) which aim to advance the development of intelligent, interactive AI agents and provide essential resources for further research.

\end{abstract}

\section{Introduction}
\label{sec:introduction}
% \todo{
% \begin{itemize}
%     \item Refer to the following paper as the base: \\
%     \href{https://arxiv.org/pdf/2305.10783}{Transforming human-centered ai collaboration: Redefining embodied agents capabilities through interactive grounded language instructions}
%     \item \href{https://www.aicrowd.com/challenges/neurips-2022-iglu-challenge/problems/neurips-2022-iglu-challenge-nlp-task} {AI Crowd Gitlab}
%     \item \href{https://github.com/microsoft/iglu-datasets} {IGLU-dataset Github}
%     \item \href{https://github.com/microsoft/greenlands}  {Greenlands Platform}
%     \item \href{https://github.com/iglu-contest/iglu-dataset-minecraft-evaluation/tree/main/minecraft_evaluation} {current mturk human eval collection}
%     \item Final list of Repos: Greenlands, dataset, and Mturk(data collection and human eval)
% \end{itemize}}

One of the enduring goals of artificially intelligent (AI) agents~\cite{winograd1972understanding} is to seamlessly interact with humans using natural language. This capability allows AI agents to learn new skills~\cite{narayan-chen-etal-2019-collaborative,zhang-etal-2021-learning,wang2023voyager} or assist in solving tasks~\cite{shridhar2020alfred,kiseleva2016predicting, li2021deus}. To achieve this, AI agents must be able to comprehend~\cite{mehta2023improving,mehta2019improving} and respond to human language, executing instructions across various environments~\cite{skrynnik2022learning}. Over the years, researchers have developed numerous tasks to address this challenge, often focusing on scenarios where humans provide instructions to achieve specific goals~\cite{gluck2018interactive,shridhar2020alfred}.
For example, in the blocks world task, the agent must understand human instructions to move blocks on a grid~\cite{winograd1972understanding,bisk2016natural}. Other setups use Minecraft~\cite{gray_craftassist_2019, fanminedojo} for tasks such as moving objects~\cite{abramson2020imitating}, simulating human behavior~\cite{park2023generative}, or performing household tasks~\cite{shridhar2020alfred, wang2023holoassist}. However, human instructions are often inherently ambiguous. To complete these tasks successfully, agents need to engage in conversations by asking clarifying questions~\cite{aliannejadi-etal-2021-building,shi2022learning,press2022measuring}, thereby creating a more user-friendly interface~\cite{nass2000machines}.

To advance and emphasize this objective of interaction-driven agent building, we organized the Interactive Grounded Language Understanding (IGLU) competition at NeurIPS in 2021\cite{kiseleva2022interactive} and 2022\cite{kiseleva2022iglu}. 
The primary aim of this competition was to foster the development of interactive agents capable of comprehending and executing grounded natural language instructions, particularly emphasizing the nuances of natural language dialogues and clarifications. \emph{The overarching goal of IGLU is to equip researchers with the data, tools, and insights necessary to evaluate the efficacy of interactive multi-turn communication with humans.}
%%% highlight as challenges. 
The first significant challenge hindering the exploration of building interactive agents is the scarcity of appropriate datasets. Moreover, the data collection process is time-consuming and difficult to set up, requiring scalable, flexible, and easily extendable data collection tools.
Another crucial requirement is an effective evaluation process and platform. Given the nature of the problem under consideration, an interactive and open evaluation platform is needed. This interactive ``human-in-the-loop" evaluation is necessary because automatic metrics such as accuracy do not thoroughly explain the performance of agents
and may not correlate well with human preferences for answers\cite{arabzadeh2024assessing,arabzadeh2024towards}. Our interactive evaluation tools provide a critical supplement to automatic evaluation metrics, providing deeper qualitative insights and ensuring the robustness and validity of the evaluation process. Such an evaluation platform also addresses concerns around data leakage from benchmark datasets into training data, as highlighted in some recent studies~\cite{balloccu-etal-2024-leak}. Finally, after running this competition for two years, the task's complexity is evident from the scores lacking in both offline and human evaluations of the agents. This emphasizes the need to release the dataset and tools to enable further research in this direction.

IDAT (IGLU Dataset And Toolkit) aims to address these challenges by making the following contributions:

%Developing interactive evaluation platforms is especially important given concerns around data leakage from benchmark datasets into training data, as highlighted in some recent studies~\cite{balloccu-etal-2024-leak}. Thus, interactive evaluation tools provide a critical supplement to automatic evaluation metrics, helping to ensure the robustness and validity of the evaluation process.
%Furthermore, it has been noted that human preferences for specific answers do not necessarily correlate with accuracy~\cite{arabzadeh2024assessing,arabzadeh2024towards}. Despite achieving high scores in offline settings, our findings reveal that current agents still fall short of meeting human expectations. Another significant challenge hindering the exploration of building interactive agents is the scarcity of appropriate datasets and scalable and easily extendable data collection tools. In response to these challenges and to encourage further research to develop intelligent interactive agents that can understand natural instructions, our work makes the following contributions:
\begin{enumerate}[leftmargin=*, nosep, label=\textbf{C\arabic*}]
    \item \textbf{Data Collection Tool:} A scalable tool designed for efficiently gathering interactive grounded language instructions and clarifying questions within a Minecraft-like, voxel world environment that can be run in a web browser, making it accessible to a large number of annotators in a crowdsourcing platform (Sec.~\ref{sec:data-collection-tool}). This tool also offers a high degree of extensibility, enabling researchers to expand existing datasets and collect more data in a customized setting. %\todo{add a line saying how it's flexible and extendable. how can other people use it.}
    \item \textbf{Multi-Modal Dataset:} 
    Based on the building structures task in a 3D voxel world, the dataset includes around $9,000$ natural language utterances, consisting of instructions given by annotators to build a structure followed by the corresponding world states, actions performed by the annotators, as well as images of the voxel world. Additionally, the datasets contain $1,182$ clarification questions posed by builders when instructions are ambiguous (Sec.~\ref{sec:datasets}). 
    \item \textbf{Human-in-the-Loop Interactive Evaluation Platform:} An interactive platform that facilitates human multi-turn communication with reinforcement learning (RL) agents by allowing annotators to compare the performance of multiple agents and providing additional qualitative analysis into their performance, thus leading to new insights into the interactive evaluation process. We released a dataset consisting of $45$ pairs of comparison games (Sec.~\ref{sec:evaluation}).
\end{enumerate}

The corpus collected using our data collection tool was leveraged and deployed during the competition, with over 55 teams utilizing it. This adoption highlights the utility of the dataset and corresponding tools in enabling research on the development of intelligent interactive agents.
All of the above resources are publicly available under the MIT license in our repositories: \textit{datasets} \footnote{\href{https://github.com/microsoft/iglu-datasets}{https://github.com/microsoft/iglu-datasets}}, \textit{data collection tool} \footnote{\href{https://github.com/iglu-contest/dataset-collection-and-evaluation}{https://github.com/iglu-contest/dataset-collection-and-evaluation}} and \textit{human-in-the-loop evaluation platform}\footnote{\href{https://github.com/microsoft/greenlands}{https://github.com/microsoft/greenlands}}. By sharing these resources with the community, our aim is to facilitate further advances in research and development, fostering the creation of more capable and interactive AI agents in a transparent manner.
%\todo{provide overview of the paper}

\vspace{-1em}
\section{Interactive Grounded Language Understanding (IGLU) Setup}
\vspace{-1em}
\label{sec:iglu-competition}

\begin{figure}
   \centering
   \vspace{-1em}
    \includegraphics[width=0.62\textwidth]{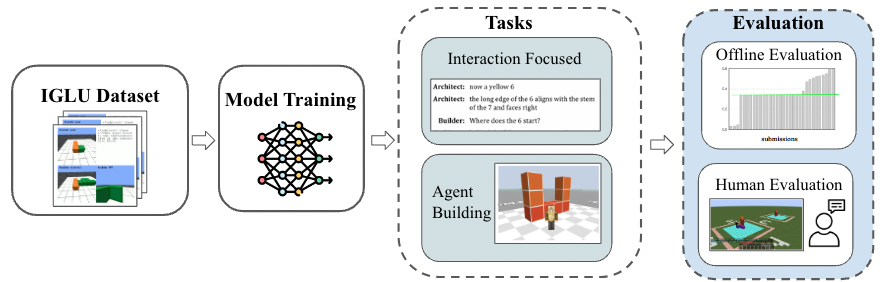}
   \caption{Interactive Grounded Language Understanding (IGLU) Setup}
       \vspace{-1em}
   \label{fig:IGLU-competition-overview}
\end{figure}

The IGLU competitions in 2021~\cite{kiseleva2022interactive} and 2022~\cite{kiseleva2022iglu} address the challenge of developing interactive agents capable of learning to solve building tasks through grounded natural language instructions in a collaborative environment. An \emph{interactive agent} is defined as one that accurately follows instructions, requests clarification when necessary, and swiftly adapts to newly acquired skills.

To approximate this scenario and simplify the study to obtain easily interpretable findings, allowing us to understand general principles, we propose the following simulated setup: The \textit{architect} and \textit{builder} communicate via a chat interface in 3D \textbf{environment}. The \textit{architect} provides the \textit{builder} with grounded instructions on constructing the target structure. The \textit{builder} may either seek clarification if the instructions are incomplete or ambiguous or proceed to execute the instructions. The \textit{architect} has the capability to observe the \textit{builder}'s actions.

To broaden the participation, the competition included two tasks: (a) \emph{an Interaction Focused Task}, and (b) \emph{an Agent Building Task}. This task setup inspired by a collaborative building task by \citeauthor{narayan-chen-etal-2019-collaborative} involves interactions between architects and builders to build a structure (Fig.~\ref{fig:iglu-mturk-pipeline}).
%Participants use the IGLU-Corpus (detailed in Sec.~\ref{sec:datasets}) to complete these tasks.

\textbf{Interaction Focused Task}
Inspired by previous works on agents seeking clarification~\cite{aliannejadi-etal-2021-building, dalton2020proceedings}, we split it into the following research questions:

\begin{enumerate}[leftmargin=*, nosep, label=\textbf{RQ\arabic*}]
    \item \textbf{When} to ask a clarifying question?
    \\ Given an instruction from the architect, a model needs to predict whether the instruction is sufficient to complete the described task or if further clarification is needed.
    \item \textbf{What} clarifying question to ask?
    \\ If the given instruction from the architect is ambiguous, a clarifying question should be raised. %This research question focuses on determining "what to ask" to clarify the given instruction. 
    %The original instruction and its clarification can then be used as input for the Builder to execute the task.
\end{enumerate}

\textbf{Agent Building Task}
This task involved building agents that take the instructions and use them to navigate and place colored blocks within the building area from a first-person perspective. The RL agent receives a score reflecting the degree of completeness of the constructed structure compared to the ground truth target structure.

\section{Data Collection Tool}
\label{sec:data-collection-tool}

We developed a scalable open-source data collection tool\footnote{\href{https://github.com/iglu-contest/dataset-collection-and-evaluation}{https://github.com/iglu-contest/dataset-collection-and-evaluation}} to facilitate the collection of multi-modal corpora (Sec.~\ref{sec:datasets}) for the collaborative building task~\cite{narayan-chen-etal-2019-collaborative, jayannavar2020learning} using the setup described in Sec.~\ref{sec:iglu-competition}
%This task involves training interactive agents to solve complex tasks while receiving natural language instructions in a Minecraft-based environment.
Unlike the data collection environment established by \cite{narayan-chen-etal-2019-collaborative}, which utilizes the Malmo platform and requires a Minecraft game server~\cite{johnson2016malmo}, our tool is entirely developed in JavaScript. This approach eliminates the need to set up a Minecraft game server, significantly simplifying the process. Additionally, our tool is highly scalable, allowing for efficient expansion and integration with crowdsourcing platforms such as Amazon MTurk.
Our data collection tool can be used to easily collect more data.

\begin{figure}[!t]
    \centering
    %\vspace{-2em}
    \begin{subfigure}[b]{0.25\textwidth}
        \centering
        \includegraphics[clip, scale=0.22]{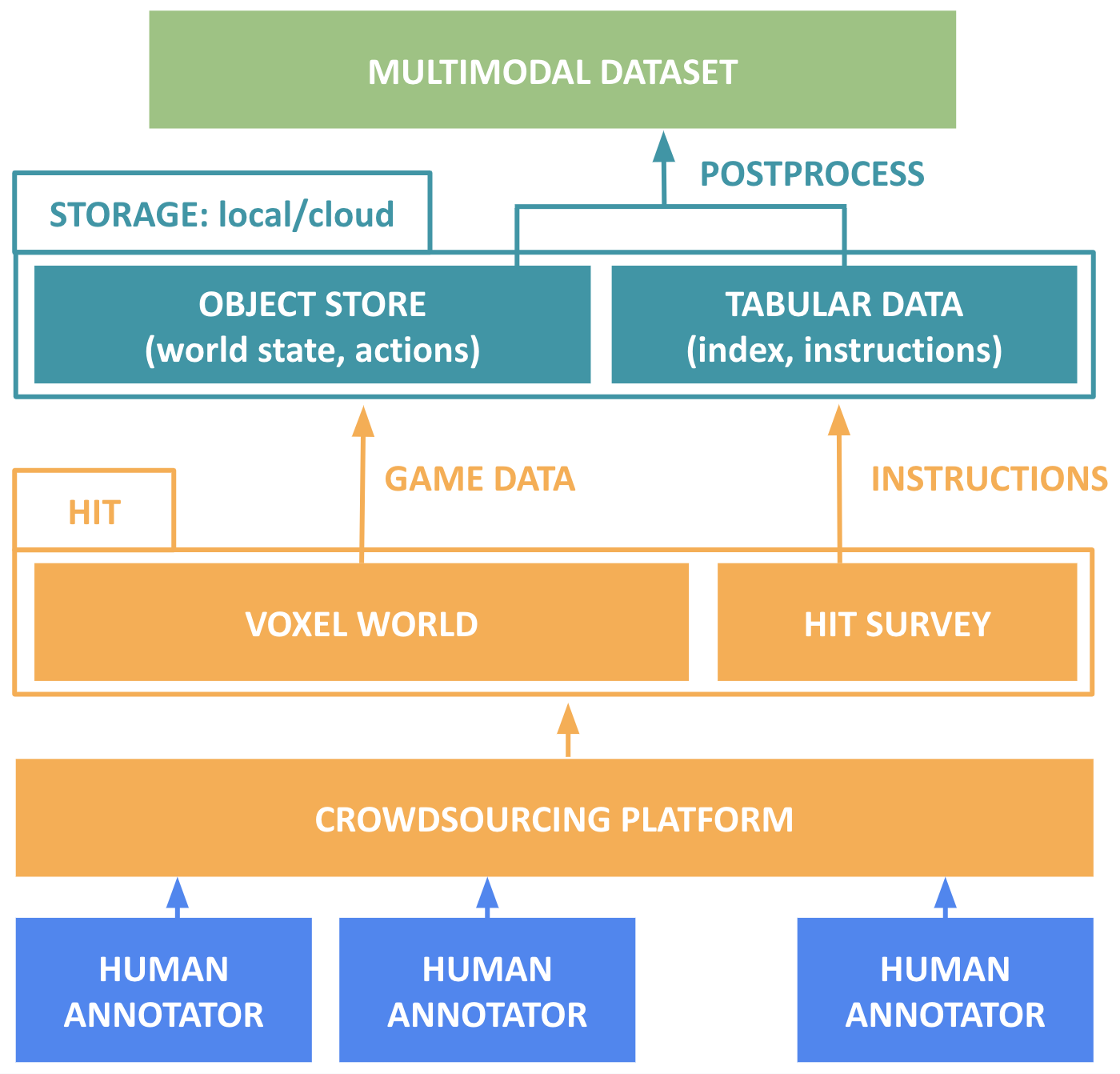}
        \caption{}
        %\vspace{-1em}
        \label{fig:iglu-mturk-pipeline}
    \end{subfigure}
    \hfill
    \begin{subfigure}[b]{0.6\textwidth}
        \centering
        %\vspace{-1em}
        \includegraphics[clip, trim=4.4cm 2.2cm 4.3cm 1.3cm, scale=0.52]{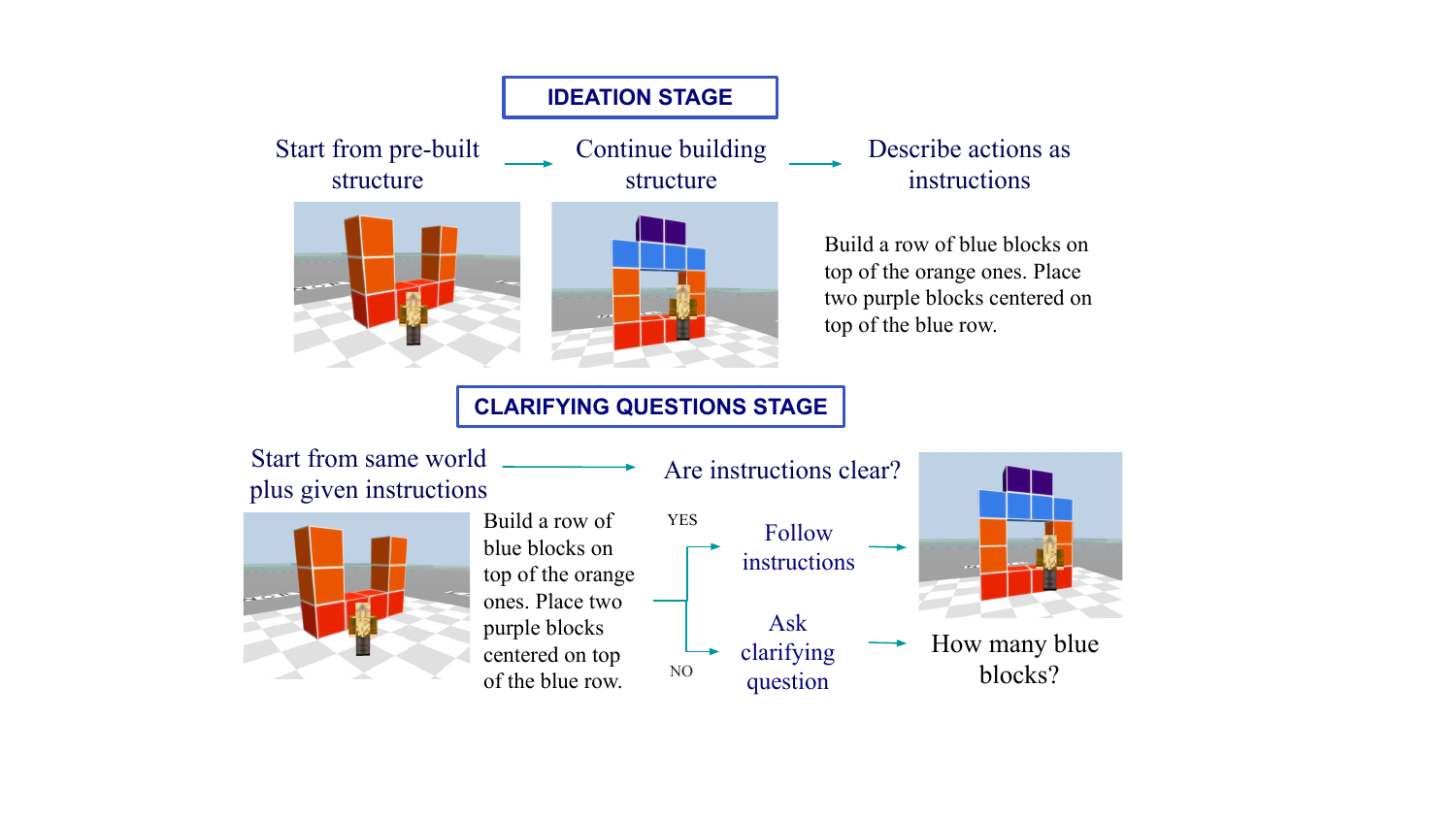}
        \caption{}
        %\vspace{-1em}
        \label{fig:nlp-task}
    \end{subfigure}
    \caption{(a) The architecture of the data collection tool. (b) The IGLU dataset collection pipeline.}
    \vspace{-1em}

    \label{fig:combined-figures}
\end{figure}

\textbf{Voxel World Environment}
\label{subsec:data-collection-tool:environment}
We harnessed a Minecraft-like game environment called CraftAssist voxel world~\cite{gray_craftassist_2019,sun2022many}
for our data collection tool which provides an immersive platform for agents to learn from language instructions and engage in fundamental navigation and building tasks, driven by its unique physics characteristics and its 3D world representation.
In the CraftAssist voxel grid world agents perform building actions within a 
$11 \times 11 \times 9$ sized build region~\cite{narayan-chen-etal-2019-collaborative} that can be recorded as action states and retrieved for future sessions. The integrated CraftAssist library supports actions such as picking, placing, and removing blocks of different colors within the voxel world. Additionally, agents can jump to place blocks, enabling the creation of structures with varying complexity. This approach ensures scalability and facilitates extensive experimentation and development within the platform. Fig.~\ref{fig:nlp-task} gives the visualization of the voxel world environment in our platform.

To reduce user friction in giving and comprehending instructions, we embedded a compass on the ground of the voxel world to aid users in understanding spatial orientations. Then in the architect task, we ask the builder to explicitly specify the view of the current structure on which the instruction is based from one of the five orientations: northward, southward, eastward, westward, or from top. Later in the builder task, we put the builder in the same orientation before providing the instructions from the architect. In this way, the architect and the builder are able to establish a shared understanding of the spatial attribute of the target structure in a multi-turn manner asynchronously.
For each task, we record the following information: gameId, stepId, and avatarInfo. avatarInfo contains the agent's spatial coordinates (x, y, z) and its corresponding pitch and yaw angles. Additionally, for the builder agent, we record a tap of the agent's actions (movement, block placement) along with the world state changes discretely. We record the architect's instructions and the builder's clarification questions.

\textbf{Data Collection Setup}
Our tool for collaborative building tasks is designed to be scalable and easily deployable to collect large datasets efficiently. It facilitates the collection of multi-modal collaborative building tasks, seamlessly integrating with crowd-sourcing platforms for efficient participant scaling. 
%Notably, the tool eliminates the need for participants to install a local client, streamlining the data collection process. Fig.~\ref{fig:iglu-mturk-pipeline} illustrates the overall design of the tool. 
Furthermore, we enhance the data collection process by introducing \emph{asynchronous turn-taking}. This means the tool no longer relies on having the same set of annotators online throughout the game. We have implemented checks to prevent a single annotator from taking on both architect and builder roles for the same structure. Importantly, this asynchronous approach allows for the simultaneous launch of multiple structures. Annotators can work on different structures concurrently without waiting for responses, saving time and making the process scalable. To facilitate clear instruction following for annotators, we utilize cardinal directions like North, South, East, and West within the voxel world.

%In our work, we use a similar setup, designed to facilitate the collection of multi-modal collaborative building tasks, seamlessly integrating with crowd-sourcing platforms for efficient participant scaling. Notably, the tool eliminates the need for participants to install a local client, streamlining the data collection process. ~\ref{fig:iglu-data-collection-tool} illustrates the overall design of the tool. Furthermore, we have enhanced the data collection process by introducing asynchronous turn-taking. This means the tool no longer relies on having the same set of annotators online throughout the game. We have implemented checks to prevent a single annotator from taking on both architect and builder roles for the same structure. Importantly, this asynchronous approach allows for the simultaneous launch of multiple structures. Annotators can work on different structures concurrently without waiting for responses, saving time and making the process scalable. To facilitate clear instruction following for annotators, we utilize cardinal directions like North, South, East, and West within the voxel world.

%\begin{figure}[!t]
%    \centering
%     \includegraphics[clip, scale=0.36]{Figures/iglu-data-collection-tool2.png}
%    \caption{The architecture of the data crowdsourcing collection tool}
%        \vspace{-1em}
%\label{fig:iglu-data-collection-tool}
%\end{figure}

We obtained approval from the Institutional Review Board (IRB) to conduct the study. We used Amazon Mechanical Turk (MTurk) as the crowd-sourcing platform to get annotations from $230$ unique annotators who provided consent to be part of the study and were paid \$15 per hour. We did not collect any personally identifiable information as part of the study. 
Each annotator submits a task referred to as a HIT
(Human Intelligence Task). A HIT consists of the
CraftAssist voxelworld~\cite{sun2022many} described in Sec.~\ref{subsec:data-collection-tool:environment} along with a HIT survey. The HIT survey is customizable for different tasks and includes rules for a given
task, a form where instructions can be submitted,
or clarifying questions asked for the building task. %Examples of the task or HITs in MTurk along with the embedded voxel world are provided in the appendix~\ref{sec:appendix-data-schema}. 
Finally, the data is stored
in two kinds of data stores for ease of access: Tables are used to save game ids, instructions, and clarifying questions while the Object Store is used
for storing files with game world states and collected actions. Although this data collection tool is currently used for the multi-turn interactions setup we described, it can be easily customized to support other general setups to collect interaction dialogs from human annotators, actions, and world states to solve building tasks.

\section{IDAT Dataset}
\label{sec:datasets}

The IDAT dataset is a comprehensive multi-modal dataset that includes instruction utterances, voxel world states at each action, and the corresponding images. Following the previously described methodology, we provide a two-part dataset: a seed dataset and the IGLU dataset \footnote{The datasets and accompanying code for analysis and visualization are publicly available at \url{https://github.com/microsoft/iglu-datasets}}.
%
%Initially, our research focused on multi-turn interactions between an architect and a builder, who collaborated to complete goal structures. This approach was inspired by the work of \citeauthor{narayan-chen-etal-2019-collaborative} (Sec.~\ref{sec:seed-dataset}). However, we encountered challenges in scaling data collection with this setup as crowdsourced annotators engaged in the collaborative building task and issued instructions for specific goal structures. Additionally, participants struggled with the complexity of multi-turn interactions during the competition. 

%To address these issues, we streamlined our approach and expanded to a setup described in Sec.~\ref{sec:iglu-corpus}, where architects provide instructions to builders for constructing free-form structures. The free-form nature of these structures allows for greater flexibility, enabling builders to create more freely, similar to real-world scenarios in interactive building games. This streamlined methodology facilitated more efficient data collection and resulted in a larger, more diverse corpus. The world states collected in the seed dataset are subsequently used as starting points for the main IGLU dataset, ensuring a diverse range of structures being built by the architect.

\subsection{Seed Dataset}
\label{sec:seed-dataset}

The seed dataset comprises multi-turn dialog sequences aimed at collaboratively building a target structure. A complete session of dialogues to achieve the target structure is referred to as a \emph{game} as shown in Fig.~\ref{fig:multi-turn}. In each \emph{turn}, an annotator assumes the role of either the \textit{architect} or the \textit{builder}.
Architects are randomly assigned a target structure from a diverse set of structures. They provide the next step instruction for the Builder. The Builder starts from scratch at the beginning of a game or builds on intermediate results by executing the Architect's instructions. If the instruction is unclear, the Builder can pose a clarifying question.

%The goal structures used in the dataset are sourced from~\cite{narayan-chen-etal-2019-collaborative}. 

%     \begin{tabular}{lcccc}
%     \hline
%     % \cline{3-6}
    
%     \hline
%      & Total & Average \\
%     \hline
%     Target Structures & 31 &  \\
%     Games & 127 & 4.09$\pm$3.69 \\
%     Turns & 584 & 4.6$\pm$3.7 \\
%     Clarifying Questions & 67 & \\

%     \hline
% \end{tabular}

Tab.~\ref{tab:multiturn_data} shows the summary of the \textit{Seed} dataset. 31 target structures are presented to the annotators to build. We process and clean the data by filtering out missing and low-quality submissions such as very short instructions having less than five words. Finally, we have $127$ completed game sessions, with the median duration of a game being around \textit{16 minutes} and the average number of turns taken to complete a game as 14 turns. 
A game session is considered complete when the Builder completes a given target structure after interacting with and following instructions provided by the Architect. This is denoted by the Architect marking the structure as \emph{``complete"}. Across all the games, we have 811 utterances or dialog interactions between the Architect and Builder annotators. The average length of instructions provided by the Architects is around 19 words, and the number of clarifying questions asked by the Builders -- $126$. On average, 2 clarifying questions are asked per game. 

The target structures have been designed to ensure a variety of building types with varying levels of difficulty. To provide a deeper understanding of the target structures in our multi-turn dataset, we performed manual labeling on the 31 structures. The types of structures and their corresponding number of structures (in brackets) in the dataset are as follows:
\begin{inparaenum}
    \item \textit{flat [7]:} all blocks on the ground
    \item \textit{flying [27]:} there are blocks that cannot be fully added without removing some other blocks 
    \item \textit{diagonal [6]:} some blocks are adjacent diagonally
    \item \textit{tricky [6]:} some blocks are hidden or they should be placed in a specific order
    \item \textit{tall [25]:} a structure cannot be built without the agent being high enough (the placement radius is 3 blocks).
\end{inparaenum} These labels are not mutually exclusive, so one structure can belong to multiple categories. We consider different categories of structures to ensure the agent uses various skills and abilities to complete the target structures. For instance, if all the structures are flat, the agent will not learn to use other actions, such as flying. %While these categories might not have been directly used in the experiments, they were employed during data collection to promote diversity in the structures and actions performed by the agent.
This diversity is essential for training a robust and adaptable agent.

\subsection{IGLU-Dataset}
\label{sec:iglu-corpus}

% From our extensive study on Multi-Turn data collection, we identified certain challenges . To enhance the crowd-sourcing process, we decided to simplify the task.
%Our approach involved removing the added complexity of building a predefined target structure. Instead, participants were free to perform free-form building actions within the voxel world while providing instructions that should allow another worker to rebuild the same structure.
%
%This modification led to the creation of single-turn task segments, where annotators collaborated asynchronously to construct the same structure. With this adjustment, we were able to collect data at a faster pace, resulting in a larger corpus comprising natural language instructions, corresponding actions performed based on those instructions, and a set of clarifying questions. We record and save actions performed by annotators in a key-value pair format that stores the movement of the agent and positional changes of blocks within the voxel world.

The multi-turn data collection process described in the previous section is fairly complex and tricky to scale. We simplify the process to be a single turn where all required attributes are captured in one shot. We first remove the complexity of building a predefined target structure. Instead, annotators are asked to perform some free-form building actions within the voxel world, while providing instructions that should allow another annotator to rebuild the same structure. These single-turn task segments enable asynchronous collaboration between annotators. This process enables the data collection at a significantly faster pace, leading to a larger corpus comprising natural language instructions, corresponding actions performed based on those instructions, and a set of clarifying questions. We record and save actions performed by annotators in a key-value pair format that stores the movement of the agent and positional changes of blocks within the voxel world.

\begin{table*}[t]
\vspace{-1.5 em}
\centering
\begin{minipage}{0.35\linewidth}
\centering
\scalebox{0.7}{
\begin{tabular}{@{}l|l@{}}
\toprule
Target Structures          & 31           \\
Completed Games          & 127          \\
Median Dur of Completed Games          & 16 mins      \\
Avg. Turns of Completed Games &  14\\
No. Instructions         & 811          \\
Avg. Len of Instructions & 19.32 words \\
No. Clarifying Questions     & 126          \\ 
Avg. Clarifying Questions per Game     & 2          \\
\bottomrule
\end{tabular}}
\caption{\small Overview of Seed Dataset}
\vspace{-1em}

\label{tab:multiturn_data}
\end{minipage}%
\hfill
\begin{minipage}{0.6\linewidth}
\centering
\scalebox{0.75}{
\begin{tabular}{@{}llll@{}}
\toprule
\\
\multicolumn{2}{l}{Instructions (train/test)} & \multicolumn{2}{l}{Avg. Length (in words)} \\ \midrule
Total            & 8136 (6843/1293)         & Instructions                     & 18.29       \\
Clear            & 7080 (5951/1129)         & Clarifying Questions            & 12.05       \\
Ambiguous        & 1056 (892/164)           &                                  &             \\ 
or Clarifying Questions        &            &                                  &             \\ 
\bottomrule
% No. Clarifying Questions        &             &              1016                    &             \\
% \bottomrule
\end{tabular}}
\caption{\small Overview of the IGLU Dataset}
\vspace{-1em}

\label{tab:single-turn-stat2}
\end{minipage}
\vspace{-0.2cm}
\end{table*}

We utilized the Seed dataset to provide diverse starting canvases for annotators as follows:

\begin{itemize}[leftmargin=*, nosep]
    \item An annotator is assigned a world state from the Multi-Turn dataset as the starting point for their building task (Fig.~\ref{fig:nlp-task}: Ideation Stage).
    \item The annotator is prompted to perform a sequence of actions for a duration of one minute.
    \item Then, the annotator is required to describe their actions in the form of a natural language instruction.
    \item Another annotator is shown the instructions and asked to perform the steps mentioned. If the instruction is unclear, the annotator specifies it as thus and asks clarification questions (Fig.~\ref{fig:nlp-task}: Clarification Question Stage).
\end{itemize}
\noindent

%The instructors answered these clarifying questions, and the data related to these clarifying questions has also been released with this dataset.
%
Tab.~\ref{tab:single-turn-stat2} presents a summary of the IGLU dataset, which consists of 8,136 pairs of actions and instructions. We clean the collected Single-Turn dataset by filtering out low-quality samples, e.g. those with very short instructions (< 5 words) or those coming from annotators who gave low-quality instructions (e.g. providing the same instruction repeatedly). In the final set, instructions consist of on average 18 words, indicating the instructions are descriptive enough for 1-minute building actions.
%
%We processed and cleaned the collected Single-Turn dataset by following a heuristic approach, which included filtering out samples where the length of instruction was very short(less than 5 words). We also checked whether the instruction was in English and evaluated jobs to remove submissions by annotators who provided low-quality instructions, such as providing the same instruction repeatedly. On average, an instruction has 18 words, which indicates the instructions are descriptive enough for a one-minute building. % We assumed the validation set could be selected from a portion of the train set.

%
In the above process, if an annotator marks the provided instruction as ambiguous to execute, they are supposed to issue a clarifying question. Otherwise, the submission is filtered out with a warning provided to the annotator. This was to ensure that every instruction annotated as ``not clear" is accompanied by at least one clarifying question. Out of 8,136 instructions, 1,056 (12.98\%) were marked as \textit{Not Clear}, thus being ambiguous, and 7,080 (87.02\%) as \textit{Clear} instructions. Hence, we have 1,056 clarifying questions, one for each ambiguous question. The average length of clarifying questions is around 12 words. Tab.~\ref{tab:unclear-questions} in the appendix exemplifies a few instructions marked as being unclear, along with clarifying questions issued by annotators.

%In addition to the processing steps for cleaning instructions, for the clarifying questions, if an annotator marked the instruction as ambiguous to execute the instruction, they were supposed to issue a clarifying question else the submission would be filtered out with a warning provided to the annotator. This was to ensure that every instruction annotated as ``not clear" is accompanied by at least one clarifying question. Out of 8,136 instructions, 1,056 (12.98\%) were annotated as \textit{Not Clear}, thus being ambiguous, and 7,080 (87.02\%) as \textit{Clear} instructions. We have 1,056 clarifying questions, for each ambiguous question. The average length of clarifying questions is around 12 words. Tab.~\ref{tab:unclear-questions} in the appendix exemplifies a few instructions marked as being unclear, along with clarifying questions issued by annotators.

The majority of clarifying questions fall into the following categories:
\begin{inparaenum}
\item \textit{Color}: Questions clarifying the color of the blocks to be used.
\item \textit{Direction/Orientation}: Questions clarifying the direction and orientation in the world.
\item \textit{Number of blocks}: Questions that clarify the number of blocks to be placed.
\item \textit{Identifying blocks to be changed}: Questions clarifying which blocks need to be changed.
\end{inparaenum}
For deeper insight, we reassessed the annotations for 100 randomly selected instructions to gauge the level of agreement among the annotators. The agreement rate among the three annotators for these 100 instructions falls within the range interpreted as ``fair" according to the Krippendorff agreement measure. This suggests that the interpretation of ambiguous instructions can be highly subjective, which further emphasizes the complexity of such a task. While one annotator may perceive an instruction as clear, another may find it ambiguous. Furthermore, different annotators may ask different clarifying questions about the same instruction, as they may identify unclear aspects from different perspectives.

The single-turn approach offers several advantages over the sequential nature of the multi-turn process of the seed dataset, one of which is the independence of each sample, allowing for easier utilization in different tasks. Each turn can be interpreted as a complete set of information, enabling flexibility in the data collection as well as it's uses. This independence allows researchers to extract valuable insights and information from individual turns without considering the entire dialogue sequence.
Moreover, the single-turn approach allows for collecting multiple clarifying questions for each instruction augmenting the richness and diversity of the dataset, enabling a deeper understanding of the nuances and challenges in generating clarifying questions. 
Both the seed and IGLU datasets offer extensive potential for studying various research questions concerning grounded language interactive agents. These datasets can be further expanded using our data collection tool.

\section{IGLU Evaluation}
\label{sec:evaluation}
\vspace{-0.3cm}
%This section offers an overview of the evaluation protocol employed during the 2022 competition, along with the post-competition human-in-the-loop evaluation of the top three RL track solutions.

While our focus in this paper is not on the solutions or baselines presented during the competition, we note them to underscore the need for the evaluation protocol we employed during the competition. This includes the development of an online interactive human evaluation platform which is a major contribution of this work. This evaluation platform serves as a crucial supplement to offline evaluation metrics, ensuring the robustness and validity of the evaluation process of interactive agents and allowing for deeper qualitative insights.
%previous works have noted human preferences for specific answers not necessarily correlating with accuracy~\cite{arabzadeh2024assessing,arabzadeh2024towards}.

%For more technical details about the solutions, please refer to papers about the competition overview~\cite{kiseleva2022iglu, mohanty2023transforming}.

\subsection{Offline Evaluation}
\label{sec:offline-evaluation-rl-nlp}
%\statement{Interaction Focused Task Evaluation}
\textbf{Interaction Focused Task Evaluation}:

%We employ two stages to address the two main research questions on handling clarifying questions.
\begin{enumerate}[leftmargin=*, nosep, label=\textbf{RQ\arabic*}]
    \item \textbf{When}?
It is evaluated as a binary classification problem: Does the provided instruction require a clarifying question? We use the macro average $F_1$ score to evaluate classifiers based on instructions marked as unclear in the corpus, ensuring a balanced measure of both precision and recall across the two classes.

\item \textbf{What}?
It is evaluated based on the quality of selected clarifying questions for unclear cases.

\end{enumerate}

We formulate the problem of ranking a pool of clarifying questions instead of generating the questions for several reasons. Generating clarifying questions in a collaborative environment is challenging, as shown in~\cite{kiseleva2022interactive}. If clarifying questions already exist in a pool, finding the most appropriate ones becomes a more manageable task than generating them from scratch~\cite{aliannejadi-etal-2021-building}. Additionally, the evaluation of classification and ranking tasks is much more well-established compared to generation tasks, as there may be multiple correct clarifying questions for any given scenario. Therefore, ranking a pool of clarifying questions allows for better evaluation and control over the output. We assess how well the model can rank a list of human-issued clarifying questions in the corpus for a given ambiguous instruction. The model's effectiveness is measured using Mean Reciprocal Rank (MRR). The average F1 score of the top three participants for RQ1 is 0.76. For ranking clarifying questions, the top three teams achieved an average MRR of 0.58. These results indicate that significant room for improvement remains, highlighting the challenges associated with these tasks.

%\todo{Negar, %can you add the the scores for top three and the baseline for both f1 and MRR in a sentence? And cite the paper which has more in-depth solutions about them.}
%This metric is particularly suitable for scenarios where the relevance of top-ranked items is crucial, as it focuses on the position of the first relevant item in the ranking.

\textbf{Agent Building Task Evaluation} 
To evaluate a RL agent, the evaluation system executes two episodes for each task using a held-out test set. Each task begins with a specific initial grid configuration and a designated target grid. The primary evaluation metric is the $F_1$ score, computed as in Algorithm~\ref{algo:f1-score} (in appendix). This score is derived by comparing the predicted modifications—differences between the initial world and the final snapshot of the building zone—to the ground truth, which includes the required blocks to be added or removed. 
Scores for each task are computed as a weighted average, with weights based on the total number of blocks that needed modification. Participants' models are required to complete two runs per task across a total of $96$ tasks, resulting in $192$ episodes. All tasks must be completed within a $60$-minute timeframe.\footnote{The system specifications for the machine running the submissions were as follows: 1 NVIDIA T4 GPU with 16 GB of memory, 8 vCPUs, 56 GB of RAM.} %The detailed results of the evaluation and descriptions of the winning solutions can be found in the appendix~\ref{appendix:rl-track-solutions}.

%In this task, agents build structures within a Minecraft-like environment from a first-person perspective, guided by instructions to navigate and place colored blocks. The agent uses multimodal observations, including both context utterances, which detail previously placed blocks, and current utterances, which describe the blocks still to be placed. The conversation between the Architect and the Builder provides these instructions. Agents also utilize an inventory that outlines the remaining blocks, a compass for precise navigation, and a Point of View (POV) that consists of an RGB image capturing the environment from the agent's perspective. The RL agent's performance is evaluated at the end of each episode using $F_1$ score described above. 

% \end{itemize}

\subsection{Human-in-the-Loop Interactive Online Evaluation: Greenlands Platform} 
\label{sec:greenlands-platform}

To facilitate the evaluation of the RL agents by human participants we developed the interactive evaluation platform. Greenlands\footnote{\url{https://github.com/microsoft/greenlands}} host agents on a Minecraft server, enabling human evaluators, sourced from a crowdsourcing platform (Amazon MTurk), to interact with and assess the agents' performance in a real time. Our findings suggest that while current RL agents exhibit a degree of functionality, they fall short of human expectations in terms of interactivity and reliability. Technical design of the platform's is provided in the appendix~\ref{sec:greenlands-platform-details-supplementary}.
\begin{table}[t]
\centering
\vspace{-2em}
\scalebox{0.82}{

\begin{tabular}{c|c|cccc}

\textbf{Agent} & \textbf{Total Games} & \textbf{Total Wins} & \textbf{Total Losses} & \textbf{Wins Against} & \textbf{Losses Against} \\ 
\hline
B & 30 & 17 (56.67\%) & 13 (43.33\%) & \begin{tabular}{@{}c@{}} MHB: 7 (53.85\%) \\ P: 10 (58.82\%) \end{tabular} & \begin{tabular}{@{}c@{}} MHB: 6 (46.15\%) \\ P: 7 (41.18\%) \end{tabular} \\ 
\hline
MHB & 28 & 15 (53.57\%) & 13 (46.43\%) & \begin{tabular}{@{}c@{}} B: 6 (46.15\%) \\ P: 9 (60.00\%) \end{tabular} & \begin{tabular}{@{}c@{}} B: 7 (53.85\%) \\ P: 6 (40.00\%) \end{tabular} \\ 
\hline
P & 32 & 13 (40.62\%) & 19 (59.38\%) & \begin{tabular}{@{}c@{}} B: 7 (41.18\%) \\ MHB: 6 (40.00\%) \end{tabular} & \begin{tabular}{@{}c@{}} MHB: 9 (60.00\%) \\ B: 10 (58.82\%) \end{tabular} \\ 

\end{tabular}}
\caption{Human evaluation results for top 3 performing agents.}
\vspace{-2em}
\label{tab:human-evaluation-results}
\end{table}
Our evaluation is focused on IGLU 2022 the top agents~\cite{kiseleva2022iglu} (\hyperlink{https://www.aicrowd.com/challenges/neurips-2022-iglu-challenge/problems/neurips-2022-iglu-challenge-rl-task/submissions/200303}{\textbf{Brain Agent (B)}} and \hyperlink{https://www.aicrowd.com/challenges/neurips-2022-iglu-challenge/problems/neurips-2022-iglu-challenge-rl-task/submissions/199644}{\textbf{MHB-Pegasus} (P)}), and  baseline model developed by IGLU team to serve as a control (\hyperlink{https://www.aicrowd.com/challenges/neurips-2022-iglu-challenge/problems/neurips-2022-iglu-challenge-rl-task/submissions/198866}{\textbf{MHB}}) ~\cite{skrynnik2022learning}, which archived the following $F_1$ scores in the offline evaluation:  
\begin{inparaenum}[(1)]
    \item \textbf{B} — $0.254$,
    \item \textbf{P} — $0.178$,
    \item \textbf{MHB} — $0.150$.
\end{inparaenum}

Our human evaluation protocol involved participants playing two separate games of interactive collaborative building task, each featuring a different agent in random order. After interacting with both agents, participants were asked to identify which agent they perceived as superior and to provide qualitative feedback on each agent's behavior. This comparative approach mitigates the inherent subjectivity by focusing on the relative performance. Participants were blinded to the identity of the agents, anonymized as \textit{Agent 1} and \textit{Agent 2}\footnote{\url{https://github.com/iglu-contest/dataset-collection-and-evaluation}}.
To ensure a fair comparison, both games assigned to a participant within a single MTurk \textit{hit} involved the same task, with identical initial and target structures. These tasks were randomly selected from our test set.

\subsubsection{Human Evaluation Results and Discussion}
\label{sec:greenlands-human-evaluation-results}

% ----------
% data from blob table:

% We had a total of 90 games. 32 games for Felipe, 30 for Happy, and 28 for MHB.

% We had only 12 qualified hits (whatever that means). We had 100 partial hits. And 68 explicitly not classified. Not sure what this means though.

% There were a total of 211 hits, but only 45 have 2 game played in them, 166 have no games recorded. There is no hit that has only one game, interestingly (this is weird). Might be too early to say but maybe the hits that have no games is because the agents just didn't work or crashed and no-one restarted them (was someone even monitoring them?). Too bad we didn't store the dates.

% Would be nice to do some semantic analysis. Perhaps a tree-analysis as well, showing the most common starting words and their subsequent words (many of the human evaluations are the same). But I don't think we have time.

% -----

We recorded a total of $45$ MTurk assignments. The human evaluations, summarized in Tab.\ref{tab:human-evaluation-results}, suggest a correlation between human preferences and offline evaluation scores, with \textit{Brain Agent} generally preferred over \textit{MHB-Pegasus}. 
%However, due to the limited sample size, the generalizability of these results may be constrained. 
However, the generalizability of these results may be limited.
Examples of human feedback on the performance of each agent are provided in appendix~\ref{sec:greenlands-human-annotations-examples-supplementary}.

Upon reviewing the qualitative feedback, we consistently see that none of the agents met human expectations or completed the tasks. Through our analysis, we identified three predominant concerns across all agents, as reported by the participants: responsiveness to commands, precision in executing actions, and compliance with given instructions.

%agents' offline scores were moderately acceptable, their effectiveness appeared to diminish further in the context of human interaction. One hypothesis is that the task setup and design may have led to a discrepancy between the \textit{shape} of the instructions provided by the participants and those in the training data. 

% andrea: the above should be substantiated by distance analysis between utterances for the same structure. But we don't have time, so leaving it open for future research.

Aligning training scenarios with the complexities of the real world is a challenging problem for interactive agents. This difficulty is evident in both offline and online evaluations of the agents. Interestingly, despite the agents' generally poor performance, there was a discernible alignment between human preferences and the outcomes of offline evaluations. This suggests that even in the presence of task completion deficits, the behavioral patterns exhibited by agents can significantly influence human perceptions of their capabilities. 
% In scenarios with ambiguous objectives, the agent's behavioral patterns and decision-making processes can have a profound impact on human judgments of its efficacy, often rivaling the importance of the final outcome.

However, the offline F1 score metric fails to reliably identify specific issues affecting the agent's performance. Additionally, human evaluators tend to provide specific instructions, especially when correcting the agents' actions, introducing a level of complexity that the metrics used in offline evaluations, which do not account for shifts or translations, fail to capture.
These findings highlight the importance of integrating human evaluations into the development cycle of interactive agents. They highlight the need for an approach that considers not only an agent's task performance but also its behavioral interactions, as both are integral to the human experience. This emphasizes the necessity of a dynamic evaluation environment and the definition of multi-dimensional utilities to gain a deeper understanding of agent systems, which cannot be fully captured through single offline metrics.
Future studies should incorporate more granular response options to capture a comprehensive range of human feedback such as allowing evaluators to express a neutral stance when no clear preference.

% itself revealed a slight (but important) shortcoming in its design: the lack of an option for 

%existed may have introduced noise into the data, potentially skewing the results. 

\section{Related Work}
\label{sec:rel-work}
\vspace{-0.2cm}
\textbf{Evolution of NLIs and Applications}Early work in Natural Language Interfaces (NLIs)~\cite{woods1972lunar, codd1974seven, hendrix1978developing} laid the foundation for understanding and designing effective interfaces for human language communication with computers. In recent years, there has been a resurgence of interest in NLIs due to advances in language understanding capabilities driven by large-scale deep learning models~\cite{devlin2018bert, LiuRoberta_2019, clark2020electra, adiwardana2020towards, roller2020recipes, brown2020language, 2303.08774,chowdhery2022palm} and the increasing demand for various applications such as virtual assistants, dialog systems~\cite{li2019dialogue,li2020guided, burtsev2017search,li-etal-2020-rethinking, li2021improving}, and question answering systems~\cite{liu2017iterative, liu2018adversarial, dinan2020second, zhang2019dialogpt}. NLIs now extend beyond traditional databases to encompass knowledge bases~\cite{copestake1990natural, berant2013semantic} to robots~\cite{tellex2011understanding}, personal assistants~\cite{kiseleva2016understanding, kiseleva2016predicting}, and other forms of interaction~\cite{fast2018iris, desai2016program, young2013pomdp,su2017building}.
\noindent
\textbf{Agent Interactivity and Learning}
The focus has shifted towards interactivity and continuous learning~\cite{mohanty2023transforming, kiseleva2014modelling}, enabling agents to interact with users~\cite{wu2023autogen}, learning new tasks from instructions~\cite{li-etal-2020-interactive, mehta2023improving,team2021creating}, assessing their uncertainty~\cite{yao-etal-2019-model}, asking clarifying questions~\cite{Aliannejadi_convAI3, aliannejadi-etal-2021-building,arabzadeh2022unsupervised,arabzadeh2022preme}, and leveraging feedback from humans to correct mistakes~\cite{elgohary-etal-2020-speak, nguyen2022framework, nguyen2019help, milani2024bedd}. Currently, LLMs are also being studied to asses uncertainty and their own errors~\cite{press2022measuring, ren2023robots}. Newer directions are studying ways of identifying possible multi-modal utility of agentic systems to~\cite{arabzadeh2024assessing,arabzadeh2024towards,pramanick2022doro}.
\noindent
\textbf{Grounded Language Understanding}
This paper focuses on grounded language understanding—connecting natural language instructions with real-world or simulated environment context and taking corresponding actions~\cite{DBLP:journals/corr/HermannHGWFSSCJ17, mitsuda-etal-2022-dialogue, ma2022dorothie}. This is crucial to enabling more effective communication between humans and intelligent agents. Our work focuses specifically on tackling grounded language understanding in the context of collaborative building tasks performed by agents~\cite{carta2023grounding, mohanty2022collecting,skrynnik2022learning}.
\noindent

\textbf{Leveraging Minecraft} 
We select Minecraft for grounded language understanding due to its distinct advantages. \citet{szlam_why_2019} highlights the benefits of an open interactive assistant in Minecraft. The game's 3D voxel grid world and adherence to simple physics rules provide ample research scenarios for reinforcement learning experimentation \cite{DBLP:journals/corr/HermannHGWFSSCJ17}. Minecraft's interactive nature, player interactions, and dialog exchanges offer diverse opportunities for grounded natural language understanding~\cite{yao2020imitation, srinet-etal-2020-craftassist,narayan-chen-etal-2019-collaborative}. The game's immense popularity ensures enthusiastic player interaction, facilitating rich human-in-the-loop studies. Minecraft's advantage extends to the availability of the highly developed set of tools for logging agents interactions and deploying agents for evaluation with human-in-the-loop, including \textit{Malmo}~\cite{johnson2016malmo}, \textit{Craftassist}~\cite{gray_craftassist_2019}, \textit{TaskWorldMod}~\cite{ogawa-etal-2020-gamification}, \textit{MC-Saar-Instruct}~\cite{kohn2020mc} and \textit{IGLU GridWorld}~\cite{zholus2022iglu}.
Among the Minecraft-based related works, MineDojo \cite{fanminedojo} is similar to IGLU in the sense that both are designed to develop intelligent agents within the expansive Minecraft environment. While MineDojo aims to build versatile agents capable of performing diverse tasks through an internet-scale knowledge base, IGLU seeks to enhance interactive agents that can understand and act on grounded natural language instructions, with a strong emphasis on natural language dialogue and clarification.
An extensive review and comparison of relevant platforms are provided in the appendix Tab.~\ref{relatedwork}.

%\todo{Negar,another paper suggested by a reviewer – DoRO: Disambiguation of Referred Object for Embodied Agents Towards Open-World Interactive Disambiguation for Robotic Grasping}

\section{Conclusion}
\label{sec:conclusions}
In conclusion, we introduce IDAT comprising the dataset, tools, and evaluation platform tailored for the development of interaction-driven agents. The dataset comprises approximately 9,000 instructions and over 1,000 clarifying questions, along with corresponding actions and grid world states for interactive building tasks in a Minecraft-like environment. The released data collection tool is scalable, supports our task setup, and can be seamlessly integrated with crowdsourcing platforms. This adaptable tool enables the collection of tailored data for specific use cases, and we recommend the collection of new test datasets to address data leakage issues. Moreover, our introduction of a human-in-the-loop interactive evaluation platform provides a robust qualitative assessment of interactive agents. The efficacy of these resources was demonstrated through the NeurIPS IGLU competition, where interactive agents learned from natural language instructions. All resources, including the dataset, data collection tool, and evaluation platform, are publicly accessible to support future research endeavors.

The complexity of the task is highlighted by the low scores observed in both offline and human evaluations of the agents. The emergence of large language and multi-modal models such as GPT-4o and Gemini~\cite{team2023gemini} offers a promising avenue for narrowing this gap, potentially equipping agents with the capability to interpret and respond to human communication in ways that more closely mirror natural human interactions. Future research should investigate the integration of these advanced models to bolster the agents' adaptability and fluency in human-like dialogue, thereby enhancing the overall naturalness and effectiveness of these interactions.

\section{Limitations}
\label{sec:limitations}

This work focused on a single environment, Minecraft, which might not be an ideal representation of real-world environments. Although Minecraft does not perfectly replicate real-world environments, it serves as a valuable platform for training agents on fundamental tasks using natural language. This is particularly relevant given the current performance limitations observed in agent-building tasks.
%The dataset is not too big due to resource constraints. However, the developed data collection tool is designed to facilitate the efficient gathering of additional data, thereby addressing these limitations.
Some may find the scale of the dataset limiting. However, the developed data collection tool is designed to facilitate the efficient gathering of additional data, thereby addressing this limitation.
%The competition's offline evaluation focused on single-turn tasks, whereas the human evaluation involved multi-turn tasks. This discrepancy highlights a need for future research to align evaluation methods more closely with real-world interactive scenarios.

\begin{ack}

We would like to express our gratitude to the many individuals who made our work possible. Our amazing co-organizers of the competition—Milagro Teruel, Arthur Szlam, Mikhail Burtsev, Mohammad Aliannejadi, Ziming Li, Zoya Volovikoa, Aleksandr Panov, and Kavya Srinet—provided invaluable assistance and contributions that were essential in building the evaluation platform, providing feedback on the data collection platform, and organizing the competition. We are grateful to Ahmed Awadallah for their guidance and support throughout this project, and to the AICrowd team for their support in hosting the competition. We extend our thanks to the team at Microsoft, including Lars Liden, Matt Mazzola, Swadheen Shukla, Qianqian Qi, Piali Choudhury, Curtis von Veh, Sam Yeh, and Jianfeng Gao, whose expertise and commitment were instrumental in the development of the Greenlands platform. Their collaborative spirit helped bring our vision to fruition. Our advisory board and previous co-organizers of the competition also deserve thanks for their input and advice. Finally, special thanks to Microsoft for their funding and overall support in making this project possible.

\end{ack}

\clearpage

\bibliographystyle{plainnat}
\bibliography{references}

\clearpage

\section*{Checklist}
\label{sec:checklist}
%%% BEGIN INSTRUCTIONS %%%
% The checklist follows the references.  Please
% read the checklist guidelines carefully for information on how to answer these
% questions.  For each question, change the default \answerTODO{} to \answerYes{},
% \answerNo{}, or \answerNA{}.  You are strongly encouraged to include a {\bf
% justification to your answer}, either by referencing the appropriate section of
% your paper or providing a brief inline description.  For example:
% \begin{itemize}
%   \item Did you include the license to the code and datasets? \answerYes{See Section~\ref{gen_inst}.}
%   \item Did you include the license to the code and datasets? \answerNo{The code and the data are proprietary.}
%   \item Did you include the license to the code and datasets? \answerNA{}
% \end{itemize}
% Please do not modify the questions and only use the provided macros for your
% answers.  Note that the Checklist section does not count towards the page
% limit.  In your paper, please delete this instructions block and only keep the
% Checklist section heading above along with the questions/answers below.
%%% END INSTRUCTIONS %%%

\begin{enumerate}

\item For all authors...
\begin{enumerate}
  \item Do the main claims made in the abstract and introduction accurately reflect the paper's contributions and scope?
    \answerYes{The main claims are reflected in the paper's contribution}
  \item Did you describe the limitations of your work?
    \answerYes{See Sec.~\ref{sec:limitations}}
  \item Did you discuss any potential negative societal impacts of your work?
    \answerYes{}
  \item Have you read the ethics review guidelines and ensured that your paper conforms to them?
    \answerYes{}
\end{enumerate}

\item If you are including theoretical results...
\begin{enumerate}
  \item Did you state the full set of assumptions of all theoretical results?
    \answerNA{}
	\item Did you include complete proofs of all theoretical results?
    \answerNA{}
\end{enumerate}

\item If you ran experiments (e.g. for benchmarks)...
\begin{enumerate}
  \item Did you include the code, data, and instructions needed to reproduce the main experimental results (either in the supplemental material or as a URL)?
    \answerNA{}
  \item Did you specify all the training details (e.g., data splits, hyperparameters, how they were chosen)?
    \answerNA{}
	\item Did you report error bars (e.g., with respect to the random seed after running experiments multiple times)?
    \answerNA{}
	\item Did you include the total amount of compute and the type of resources used (e.g., type of GPUs, internal cluster, or cloud provider)?
    \answerNA{}
\end{enumerate}

\item If you are using existing assets (e.g., code, data, models) or curating/releasing new assets...
\begin{enumerate}
  \item If your work uses existing assets, did you cite the creators?
    \answerYes{We have cited authors of previous relevant works through all the sections of our paper.}
  \item Did you mention the license of the assets?
    \answerYes{See Sec.~\ref{sec:introduction}}
  \item Did you include any new assets either in the supplemental material or as a URL?
    \answerYes{This paper focuses on the dataset, its analysis, and the accompanying tools. We have made all data, tools, and codes publicly available on GitHub, with direct links provided in the introduction (Sec.~\ref{sec:introduction}) and relevant sections.}
  \item Did you discuss whether and how consent was obtained from people whose data you're using/curating?
    \answerYes{We went through IRB and obtained consent from participants. See Sec.~\ref{sec:data-collection-tool}}
  \item Did you discuss whether the data you are using/curating contains personally identifiable information or offensive content?
    \answerYes{We note in Sec.~\ref{sec:data-collection-tool} that we did not collect or use any personally identifiable information.}
\end{enumerate}

\item If you used crowdsourcing or conducted research with human subjects...
\begin{enumerate}
  \item Did you include the full text of instructions given to participants and screenshots, if applicable?
    \answerYes{We have included screenshots and instructions.}
  \item Did you describe any potential participant risks, with links to Institutional Review Board (IRB) approvals, if applicable?
    \answerNA{}
  \item Did you include the estimated hourly wage paid to participants and the total amount spent on participant compensation?
    \answerYes{Please see Sec.~\ref{sec:data-collection-tool}}
\end{enumerate}

\end{enumerate}

%%%%%%%%%%%%%%%%%%%%%%%%%%%%%%%%%%%%%%%%%%%%%%%%%%%%%%%%%%%%

\appendix

\clearpage

\appendix

% \section{Appendix}
\appendix
\label{sec:appendix}

\section{Comparison between related platforms}
\label{sec:related-tab-appedix}

Tab.~\ref{relatedwork} showcases a comparison of the IGLU dataset with other related platforms across several dimensions, including dataset size, support for collaborative instructions between humans and AI, availability of data collection and training environments, and provision of a human evaluation platform. As depicted in this table, the IGLU dataset distinguishes itself by offering a comprehensive suite of features, including a relatively large dataset size, tools for collaborative interactions, accessible data collection and training environments, and a robust human-in-the-loop evaluation platform. This positions IGLU as a versatile and valuable resource in the field of interactive grounded language understanding.

\begin{table*}[ht!]
    \centering

    \scalebox{0.635}{
    \begin{tabular}{lllcccc}
        \toprule
        Dataset & Settings & Size of dataset & 
        \rotatebox{90}{\makecell{Collaborative\\instructional\\(AI/Human)}} & 
        \rotatebox{90}{\makecell{Data\\collection \\tool\\availability}} &                
        \rotatebox{90}{\makecell{Training\\ environment\\availability}} &
        \rotatebox{90}{\makecell{Human\\Evaluation\\Platform}} \\
        \midrule
        SHRDLURN\cite{DBLP:journals/corr/WangLM16} & Building game & 100 games (10,223 utterances) & \tikzcmark & \tikzxmark & \tikzcmark & \tikzxmark\\
        Voxelurn\cite{DBLP:journals/corr/WangGLM17} & Building structures& 230 structures (36,589 utterances) & \tikzcmark & \tikzxmark & \tikzcmark  & \tikzxmark\\
        CEREAL-BAR\cite{DBLP:journals/corr/abs-1910-03655} & Collaborative games & 1202 & \tikzcmark  & \tikzxmark & \tikzcmark  & \tikzxmark \\
        ALFRED\cite{shridhar2020alfred} & Household tasks & 25,743 & \tikzxmark & \tikzxmark & \tikzcmark &  \tikzxmark\\
        CVDN\cite{DBLP:journals/corr/abs-1907-04957} &  Navigation & 2050 & \tikzcmark & \tikzcmark & \tikzcmark & \tikzxmark \\
        TEACh\cite{DBLP:conf/aaai/PadmakumarTSLNG22} & Household tasks  & 3215 & \tikzcmark & \tikzxmark & \tikzcmark & \tikzxmark\\
        MineDojo \cite{fanminedojo}& Minecraft & 730K YouTube videos, 7K Wiki pages, 340K Reddit posts & \tikzcmark & N/A & N/A & \tikzxmark \\
        MineRL \cite{guss2019minerl} & Minecraft & 500 video hours & \tikzxmark & \tikzcmark & \tikzcmark & \tikzcmark \\
        HoloAssist \cite{wang2023holoassist} & Physical tasks & 166 video hours & \tikzcmark & N/A & N/A & \tikzxmark \\
        \textbf{IGLU (our work)} & \textbf{Collaborative} building & 8,947 utterances/1,182 clarifying questions & \tikzcmark & \tikzcmark & \tikzcmark & \tikzcmark  \\
        \bottomrule
    \end{tabular}
    }
    \caption{Comparison between relevant platforms.}
    \label{relatedwork}
\end{table*}

\noindent

\section{Data Collection Tool}
\label{sec:appendix-data-collection}

\begin{figure}[ht]
    \centering
        \includegraphics[scale=0.36]{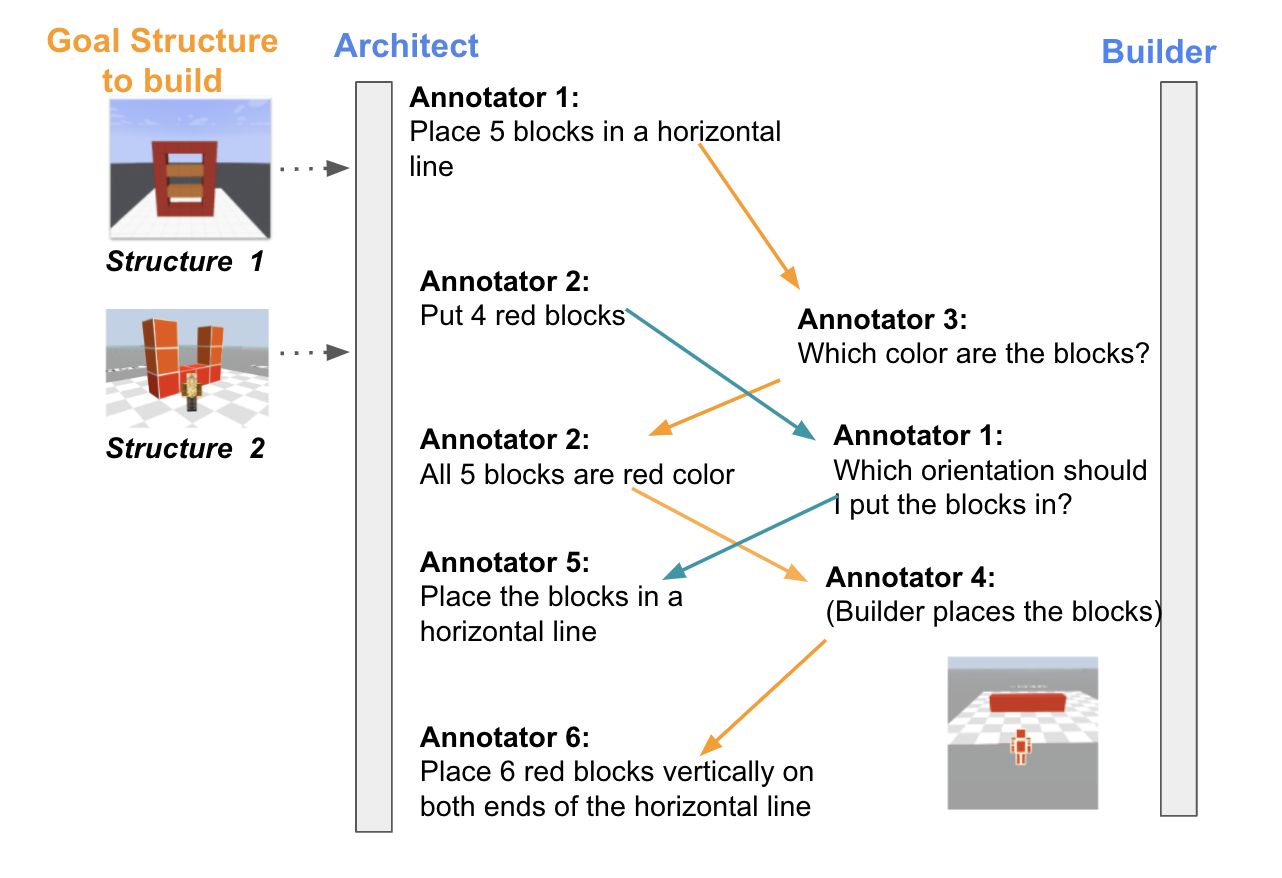}
    \caption{Example of seed data collection, where the Architect can see the goal structure and provides instructions for the Builder. The blue arrows indicate turns for the first goal structure, the orange arrows indicate turns for the second goal structure. Annotators can switch roles between architect and builder for different structures.}
    \vspace{-1em}
\label{fig:multi-turn}
\end{figure}

\textbf{Seed Data Collection:} In Figure~\ref{fig:multi-turn}, we illustrate an example of the seed multi-turn interaction data collection. In this scenario, the Architect can observe the goal structure and offer instructions to the Builder. The blue arrows represent the turns associated with the first goal structure, while the orange arrows correspond to the turns related to the second goal structure. Annotators can switch roles between architect and builder for different structures. Figure~\ref{fig:multi-turn} illustrates this concept of our data collection methodology with different annotators (1, 3, 2, 4, and 6) collaborating to construct Structure 1. Annotators can switch roles between architect and builder for different structures.

\begin{figure*}[htbp]
  \centering
  \subfloat[Architect Task in MTurk]{\includegraphics[width=0.8\textwidth]{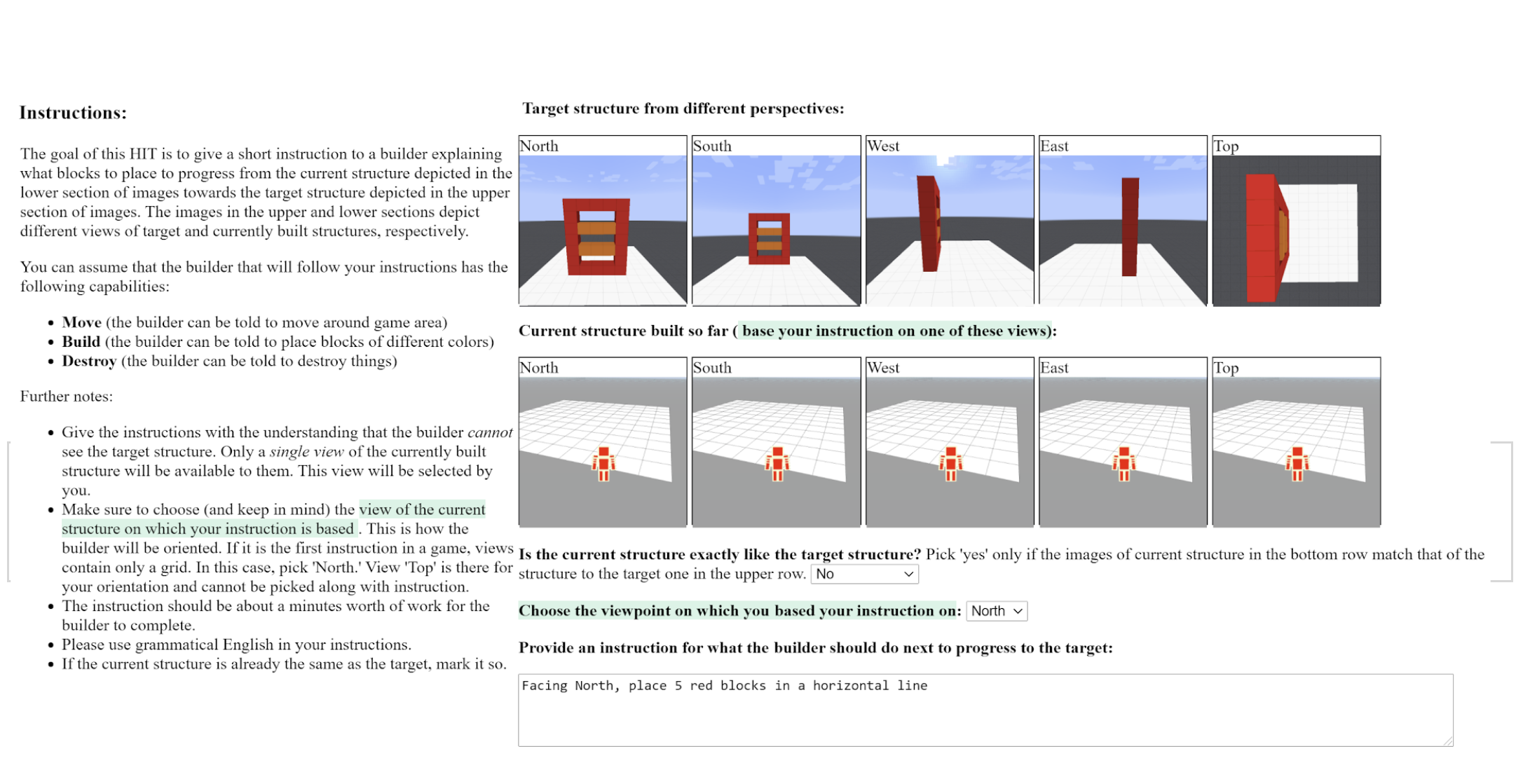}}\quad
  \subfloat[Builder Task in MTurk]{\includegraphics[width=0.75\textwidth]{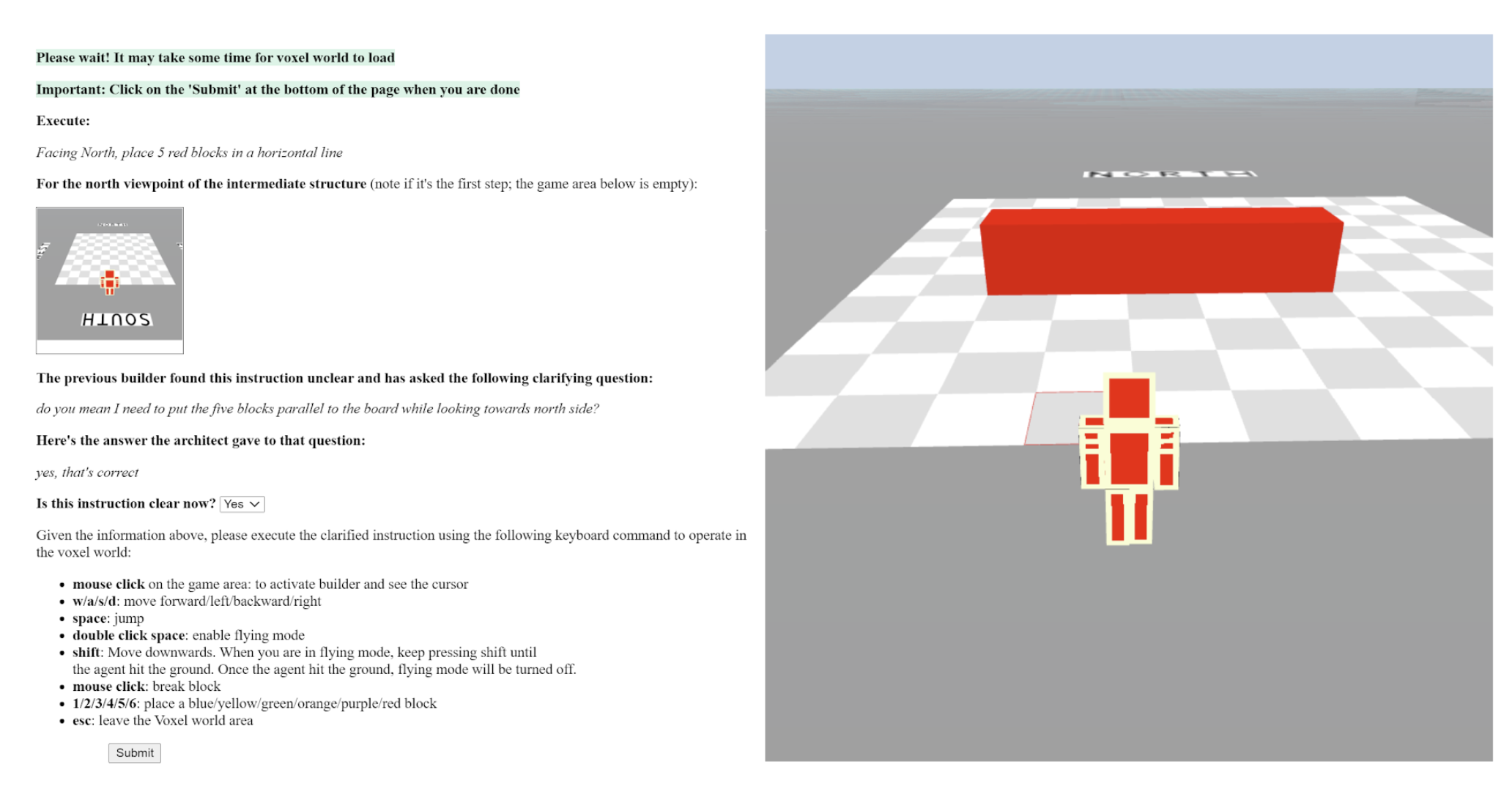}}\quad
  \subfloat[Intermediate Architect Task in MTurk]{\includegraphics[width=0.8\textwidth]{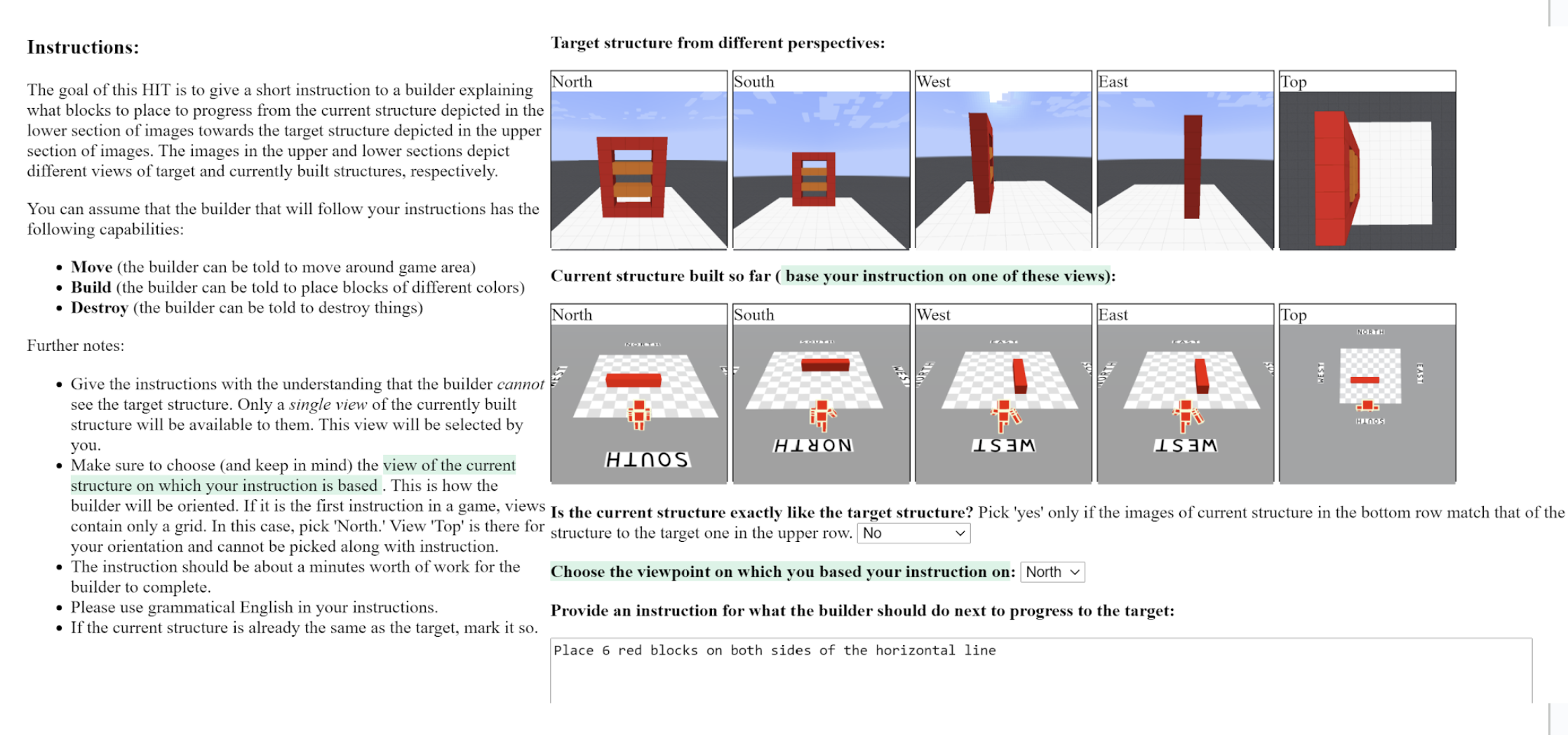}}
  \caption{MTurk view of the data collection tool. }
  \label{fig:mturk-view-tool}
\end{figure*}

% Fig.~\ref{fig:iglu-data-collection-tool} illustrates the overall design of the tool. Our tool can be integrated with crowd-sourcing platforms to provide an interface for participants to complete tasks. 
Figure~\ref{fig:mturk-view-tool} demonstrates MTurk views of the Data Collection Tool (Section~\ref{sec:data-collection-tool}) for the Seed Dataset (Section ~\ref{sec:seed-dataset}). We have the Architect Task, where the Architect provides instructions to the Builder based on the provided target structure. Next, we have the Builder Task, where instructions and the current structure built so far are shown. The Builder can mark the instructions as unclear or will follow the instructions by adjusting blocks in the voxel world. Finally, we have the Intermediate Architect Task, where the Architect is shown the progress of the structure built so far and provides the next instruction.

% \begin{figure}[!t]
%     \centering
%      \includegraphics[clip, scale=0.36]{Figures/iglu-data-collection-tool2.png}
%     \caption{The architecture of the data crowdsourcing collection tool}
%         \vspace{-1em}
% \label{fig:iglu-data-collection-tool}
% \end{figure}

\definecolor{eclipseStrings}{RGB}{42,0.0,255}
\definecolor{eclipseKeywords}{RGB}{127,0,85}
\definecolor{mylightgray}{RGB}{238,238,238}
\colorlet{numb}{magenta!60!black}

\lstdefinelanguage{json}{
    basicstyle=\normalfont\ttfamily,
    commentstyle=\color{eclipseStrings}, % style of comment
    stringstyle=\color{eclipseKeywords}, % style of strings
    numbers=left,
    numberstyle=\scriptsize,
    stepnumber=1,
    numbersep=8pt,
    showstringspaces=false,
    breaklines=true,
    frame=lines,
    backgroundcolor=\color{mylightgray},
    string=[s]{"}{"},
    comment=[l]{:\ "},
    morecomment=[l]{:"},
    literate=
        *{0}{{{\color{numb}0}}}{1}
         {1}{{{\color{numb}1}}}{1}
         {2}{{{\color{numb}2}}}{1}
         {3}{{{\color{numb}3}}}{1}
         {4}{{{\color{numb}4}}}{1}
         {5}{{{\color{numb}5}}}{1}
         {6}{{{\color{numb}6}}}{1}
         {7}{{{\color{numb}7}}}{1}
         {8}{{{\color{numb}8}}}{1}
         {9}{{{\color{numb}9}}}{1}
}

\subsection{Data Schema}
\label{sec:appendix-data-schema}
This section describes the schema of the data we collected in both the architect and builder tasks, along with a shortened version of an example data for illustration purpose.

\begin{lstlisting}[language=json,numbers=none, caption={Architect data schema}]
{
    "$id": "iglu.architect.schema.json",
    "$schema": "http://json-schema.org/draft-07/schema#",
    "title": "Data schema for architect in IGLU",
    "type": "object",
    "properties": {
        "gameId": {
            "description": "unique id for each game session (where a target strcuture is defined)",
            "type": "integer"
        },
        "stepId": {
            "description": "a monotonically increasing id, identifying which step the architect is in",
            "type": "integer"
        },
        "avatarInfo": {
            "type": "object",
            "properties": {
                "perspective": {
                    "description": "from which perspective the architect is giving command",
                    "type": "string",
                    "enum": [
                        "north",
                        "south",
                        "east",
                        "west"
                    ]
                }
            }
        },
        "command": {
            "description": "the command architect gives after he/she sees the target structure and the current world state",
            "type": "string"
        }
    }
}
\end{lstlisting}

\begin{lstlisting}[language=json,numbers=none, caption={Builder data schema}]
{
    "$id": "iglu.builder.schema.json",
    "$schema": "http://json-schema.org/draft-07/schema#",
    "title": "Data schema for builderin IGLU",
    "type": "object",
    "properties": {
        "gameId": {
            "description": "unique id for each game session (where a target strcuture is defined)",
            "type": "integer"
        },
        "stepId": {
            "description": "a monotonically increasing id, identifying which step the builder is in",
            "type": "integer"
        },
        "avatarInfo": {
            "type": "object",
            "properties": {
                "pos": {
                    "description": "an array of three floats representing avatar's position. i.e. [x, y, z]",
                    "type": "array"
                },
                "look": {
                    "description": "an array of two floats representing avatar's pitch and yaw. i.e. [pitch, yaw]",
                    "type": "array"
                }
            }
        },
        "worldEndingState": {
            "description": "the ending state of the world after builder has interact with it",
            "type": "object",
            "properties": {
                "blocks": {
                    "description": "An array of blocks info",
                    "type": "array",
                    "items": {
                        "description": "An array of four elements: [x, y, z, blockId]",
                        "type": "array"
                    }
                }
            }
        },
        "tape": {
            "description": "A string representation of the tape recording builder's interaction and world state changes, see example_data.txt",
            "type": "string"
        },
        "clarification_question": {
            "description": "The question builder asks for clarification when they feel confused about their task",
            "type": "string"
        }
    }
}
\end{lstlisting}

\begin{lstlisting}[language=json,numbers=none, caption={Example of collected data (shortened)}]
{
  "gameId": 19,
  "stepId": 1,
  "avatarInfo": {
    "pos": [
      -0.5333829883845848,
      65.07999999999996,
      -3.6806624583844014
    ],
    "look": [
      -1.0720000000000007,
      -15.771999999999965
    ]
  },
  "worldEndingState": {
    "blocks": [
        [-2, 63, 1, 50],
        [-1, 63, -2, 57],
        [-1, 63, 1, 50],
        [0, 63, -3, 57],
        [1, 63, 0, 57],
    ]
  },
  "tape": [
    "0 set_look (-0.004, 0)",
    "1 set_look (-0.044, -0.042)",
    "2 action step_backward",
    "3 pos_change (-0.10159854456559483, 63, 0.014814775657966633)",
    "4 action select_and_place_block 50 1 63 0",
    "5 block_change  (1, 63, 0, 0, 50)",
    "..."
  ],
  "clarification_question": "null"
}
\end{lstlisting}

\section{IGLU Dataset}
\label{sec:appendix-dataset}
\textbf{Examples of IGLU-Dataset:} Tab.~\ref{tab:unclear-questions} provides examples of instructions marked as unclear in the dataset along with different kinds of clarifying questions posed by annotators (Sec.\ref{sec:iglu-corpus}). Clarifying questions consist of topics such as color, direction, and identification of blocks.

\begin{table*}[ht]
\scalebox{0.75}{
\begin{tabular}{|p{11cm}|p{6cm}|}
\hline
Instruction & Clarifying Question
\\[0.5 em] \hline
Place four blocks to the east of the highest block, horizontally. & Which color blocks?\\[0.5em] \hline
Destroy 2 purple blocks and then build 3   green blocks diagonally. & Which two purple   blocks need to be destroyed? \\ \hline
Destroy the 3 stacked red blocks on the east side. Replace them with 3   stacked blue boxes & Which three of the four stacked red   blocks on the east side need to be destroyed? \\ \hline
Make a rectangle that is the width and height of the blue shape and fill it in with purple blocks. & Which side I need to make the rectangle   is not clear \\ \hline
Facing South remove the rightmost purple block. Place a row of three orange blocks to the left of the upper leftmost purple block. Place two orange blocks above and below the leftmost orange block. & Which one of the rightmost blocks should be removed? \\ \hline
Facing north and purple-green blocks will be arranged one by one. & Where would you like to place the purple and green blocks exactly? \\ \hline
\end{tabular}}
\caption{Examples of Unclear Instructions and corresponding Clarifying Questions}
\label{tab:unclear-questions}
\end{table*}

\section{IGLU-2022 Evaluation protocol}
\label{sec:appendix-iglu-protocol}
During the competition participating in the IGLU competition involves three phases:
\begin{enumerate}
    \item \textbf{Sign Up}: Participants must register on the AIcrowd website to access the competition details and starting kits.
    \item \textbf{Prepare and Train}: After registration and accepting the rules of the competition, participants can start by using prepared baselines and instructions. This involves configuring and training hybrid, RL and NLP models to interact with the IGLU environment.
    \item \textbf{Submit Models}: Once training is complete, participants submit their models to the AIcrowd for automated evaluation. The performance is assessed over a fixed number of episodes, and results are ranked on the competition leaderboard.
\end{enumerate}

\begin{algorithm}[ht!]
\caption{Computation of the $F_1$ Score}
\label{algo:f1-score}
\KwIn{$G$ -- current state of the grid, $G_0$ -- initial state of the grid, $T$, the target state of the grid.}
\KwOut{$F_1$, the computed $F_1$ score.}
\SetAlgoLined

$M \gets G - G_0$ \tcp*[r]{Compute the difference between current and initial grids}
$A \gets \text{argmax-intersection}(M, T)$ \tcp*[r]{Find indices where current grid's modifications best intersect with the target}
$I \gets \text{intersection}(M, A)$ \tcp*[r]{Calculate the number of correct modifications}
$P \gets \frac{I}{|T|}$ \tcp*[r]{Calculate precision as the ratio of correct modifications to target size}
$R \gets \frac{I}{|{i : M[i] \neq 0}|}$ \tcp*[r]{Calculate recall as the ratio of correct modifications to all modifications}
$F_1 \gets \frac{2 \cdot P \cdot R}{P + R}$ \tcp*[r]{Compute $F_1$ score, the harmonic mean of precision and recall}
\Return{$F_1$};

\end{algorithm}

\section{IGLU 2022 Winning Solutions of Agent Building Task}\label{appendix:rl-track-solutions}

Table~\ref{tab:rl-results} presents the results of the winners of the RL task and compares them with the proposed baseline. The \emph{Happy Iglu} team won by a significant margin, offering a multimodal end-to-end solution. Team \emph{FelipeB} and the \emph{Chuang} team improved the NLP part of the MHB (Multitask Hierarchical Builder) baseline to arrive at their solutions. A more comprehensive overview of the solutions is provided below.

\begin{table}[ht!]
\centering
\begin{tabular}{rllllcc}
    \toprule
    Team & Approach & $F_1$   & Precision & Recall & Ep. Length & \# of Submissions \\
    \midrule
    \emph{Happy Iglu}  & Brain Agent & 0.254       & 0.331     & 0.264  & 391        & 89         \\
    \emph{FelipeB}    & MHB-Pegasus  & 0.178       & 0.335     & 0.153  & 283        & 18         \\
    \emph{Chuang}  & MHB-Tuned     & 0.156       & 0.303     & 0.138  & 294        & 31         \\
    \midrule
    Baseline (ours) & MHB             & 0.150       & 0.256     & 0.134  & 281        & -  \\
    \bottomrule
\end{tabular}
\caption{Results of the winners of Building Task.}
\label{tab:rl-results}
\end{table}

\paragraph{First Place: \emph{Happy Iglu Team}}

The \emph{Happy Iglu Team} developed an end-to-end RL approach, called Brain Agent, to effectively address the challenges in the IGLU environment. Their approach encompassed several main strategies. Firstly, they crafted a sophisticated reward function that integrates task-specific rewards and penalties. They used the $F_1$ score for evaluation, parameter tuning, and selecting the best model during training.

The team employed advanced representation learning techniques to distill relevant information from high-dimensional inputs such as \texttt{grid} and \texttt{target\_grid} for the value function, incorporating additional features like compass orientation and color count. Information about \texttt{grid} and \texttt{target\_grid} was absent during testing but utilized in training exclusively by the critic. An auxiliary loss---a \texttt{grid} reconstruction loss---was applied to optimize state utilization, ensuring the agent properly memorized the current environmental state. To address partial observability, the processing of past observation trajectories utilized the \textsc{TrXL} transformer architecture.

Lastly, \textsc{COCO-LM-large} was utilized to generate embedding vectors for each instructional input (utterances). These findings were combined, resulting in high performance scores in the IGLU environment. The model was trained using the \textsc{Brain Agent}\footnote{\href{https://github.com/kakaobrain/brain-agent}{https://github.com/kakaobrain/brain-agent}} distributed RL framework.

\paragraph{Second Place: team \emph{FelipeB}}

This solution focused on addressing the limitations of the MHB baseline's NLP component, particularly the low performance of the T5 model, which was reflected by its low BLEU score. To solve this, the \textsc{Pegasus} model, pre-trained for summarization~\cite{zhang2020pegasus}, was chosen to translate utterances from \textit{architect} into commands. The \textsc{Pegasus-Large} model was trained using the same data augmentations as the original baseline. With careful hyperparameter selection and replacing the T5 model, the BLEU score significantly improved from $0.3$ to $0.95$, and the $F_1$ score of the entire pipeline rose from $0.15$ to $0.178$, contributing greatly to the success in the competition.

% During training, the building episodes were split to ensure that no event, regardless of color permutations, appeared in both the training and validation sets. However, the model had inconsistent BLEU score improvements. Even though augmented examples from the train set were included in the validation set, the model that resulted from the improper separation was the best performing in the online test. Finally, the best combination of training and inference history was achieved by training with a history of 10 \textit{architect} utterances and inferring with all the possible available history.

\paragraph{Third Place: Team \emph{Chuang}}

The team focused on transforming the problem of creating a voxel grid into a text-to-video task, using a temporal dimension to represent the grid's third dimension. They utilized an open-source video diffusion model, enhanced with context prompting by integrating the starting grid into each language instruction, improving the model's ability to generate the desired output. This approach applied to the IGLU task, outperformed the T5 model in local tests but faced challenges in external validation. 

\paragraph{Baseline: Multitask Hierarchical Builder}

The MHB baseline incorporates three core components to handle task execution based on given instructions:

NLP Module: Utilizes a finetuned T5 encoder-decoder transformer model to predict block coordinates and IDs based on textual instructions. This model is specifically trained on the IGLU dataset to generate sequences of building commands from dialogues. To handle changes in context, it incorporates the last few interactions during fine-tuning and inference to improve prediction accuracy.

Heuristic Module: This Python-based module processes the output from the NLP module to sequentially generate block placement or removal actions. It employs heuristics to determine the sequence of these actions, ensuring each block is handled individually, which aligns with the atomic operational nature of the subsequent RL module.

RL Module: Operates on visual input from the environment, along with data about the inventory and a target block, to execute the physical task of placing or removing a block. This module uses a convolutional ResNet architecture combined with an LSTM to integrate and process environmental data and execute actions based on a reinforcement learning policy trained with the Asynchronous PPO algorithm. A detailed overview of the baseline can be found at this link~\footnote{\href{https://gitlab.aicrowd.com/aicrowd/challenges/iglu-challenge-2022/iglu-2022-rl-mhb-baseline}{https://gitlab.aicrowd.com/aicrowd/challenges/iglu-challenge-2022/iglu-2022-rl-mhb-baseline}}.

% ========
\section{Human Evaluation Platform Details}
\label{sec:greenlands-platform-details-supplementary}

This section provides a technical overview of the \textit{Greenlands} platform. A more detailed description can be found in the project's code repository\footnote{\url{https://github.com/microsoft/greenlands/blob/main/Docs/Home.md}}.

The \textit{Greenlands} platform is an integration of three principal \textit{components}:

\begin{enumerate}[leftmargin=*, nosep, label=\textbf{C\arabic*}]
    \item \textbf{Server} — This central server operates a customized version of the standard Minecraft server, enabling human-agent interaction through specialized behaviors and commands. It is responsible for coordinating human players, pairing them with agents, and managing game progression by tracking in-game events, initializing game worlds, and monitoring the completion of games.
    \item \textbf{Service} — This is a standalone server that performs dual functions: it stores configurations for tasks designated for human evaluation and provides a user interface for competition organizers, such as those from \textit{IGLU}, to administer these tasks.
    \item \textbf{Agent Toolkit} — Acting as a Python-based wrapper for the IGLU agents, the \textit{Agent Toolkit} executes these models within their original training environment, Gridworld, capturing environmental changes and relaying them to the Minecraft server to synchronize the agent's actions with the human player's experience.
\end{enumerate}

A bi-directional communication channel facilitates the exchange of game \textit{events} between the Minecraft server and each Agent Toolkit instance. These events encompass a set of discrete actions within the game: \begin{inparaenum}[(1)]
    \item Player joining the game
    \item Chat interactions
    \item Player movements
    \item Block placements
    \item Block removals
    \item Turn endings
    \item Game conclusions
\end{inparaenum}

Incoming events are processed in the corresponding game world, either within the Minecraft server or the Agent Toolkit. Upon initialization of an Agent Toolkit instance, it attempts to connect to the Minecraft server, establishing a communication link that allows the server to recognize active agents available for gameplay.

The Agent Toolkit is designed to enable a single agent instance to concurrently participate in multiple games, assuming the agent model does not maintain an internal state between steps in the environment. For the purposes of our human evaluation, however, each Agent Toolkit instance was restricted to a single game to optimize inference speed and simplify monitoring. Although multiple agent instances were operational simultaneously as different Agent Toolkit processes.

It is important to note that the Gridworld environment, where the agents are executed, does not replicate the exact same physics as Minecraft. It also differs slightly in specific action parameters, such as the block placement/removal radius and the permissible collision boundaries. Consequently, agent physics is not applied within the Minecraft server; instead, agent actions are first simulated in the Agent Toolkit, and then the final state is mirrored in the Minecraft environment. In contrast, the human player's interactions are processed directly by the Minecraft server, with relevant state information transmitted to the Agent Toolkit so the model can consume it.

A human participant wanting to play with an agent would need to go through the following sequence:

\begin{itemize}
    \item Acquire a \textit{join code} from the competition organizers, which was created beforehand through \textit{Service}'s web interface and specifies the agent and task for the game.
    \item Connect to the \textit{Greenlands} Minecraft server endpoint, entering the Lobby World where the sole possible action is to input the \textit{Join Code}.
    \item Upon code submission, the server alerts the designated agent that a game will commence, generates a new \textit{Game World} with pre-set structures, and places the human as the \textit{architect} and the agent as the \textit{builder}. The architect has the ability to fly around the world, observing both the target structure and the agent within its \textit{build area}. The agent is confined to its built area and is unable to traverse outside of it or interact with elements beyond its designated borders.
    \item The game officially begins when the human, acting as the architect, compares the current state of the agent's build area with the target structure. The human then formulates and sends an utterance to guide the agent, who serves as the builder, towards achieving the goal. After issuing this instruction, the human ends their turn using a specific command provided by the platform. Subsequently, the agent takes its turn, receiving the current world state and the entire chat history as input. It is then instructed to perform actions until it either exceeds a predefined maximum number of steps or determines that its turn is complete.
    \item The turn-based interaction is conducted in a loop until the human player either (a) acknowledges that the agent has accurately completed the target structure, or (b) determines that the agent has reached an irrecoverable state and cannot complete the structure. At this point, the human issues an \textit{End Game} command, which includes an indication of whether the game concluded successfully or not.
    \item Subsequently, the platform dismantles the game world, readies the agent for a new game and returns the human to the Lobby World.
    \item Upon the game's conclusion, the platform dismantles the Game World and informs the corresponding Agent Toolkit instance that the session has ended, preparing it for subsequent matches. The human participant is then teleported back to the Lobby World. Additionally, the server issues a \textit{Completion Code} to the participant via the chat box. The participant will enter this code into the appropriate field of the MTurk task and let the competition organizers use it to query the \textit{Service} and retrieve the complete log of the game, including the human player's assessment of whether the game concluded successfully.
\end{itemize}

\subsection{Gameplay Screenshots}
\label{sec:greenlands-platform-screenshots-supplementary}

The following images illustrate the experience of a human participant from the moment they join the \textit{Greenlands} Minecraft server till they finish the game and obtain their \textit{confirmation code}.

\begin{figure}[H]
    \centering
    \includegraphics[width=0.8\textwidth]{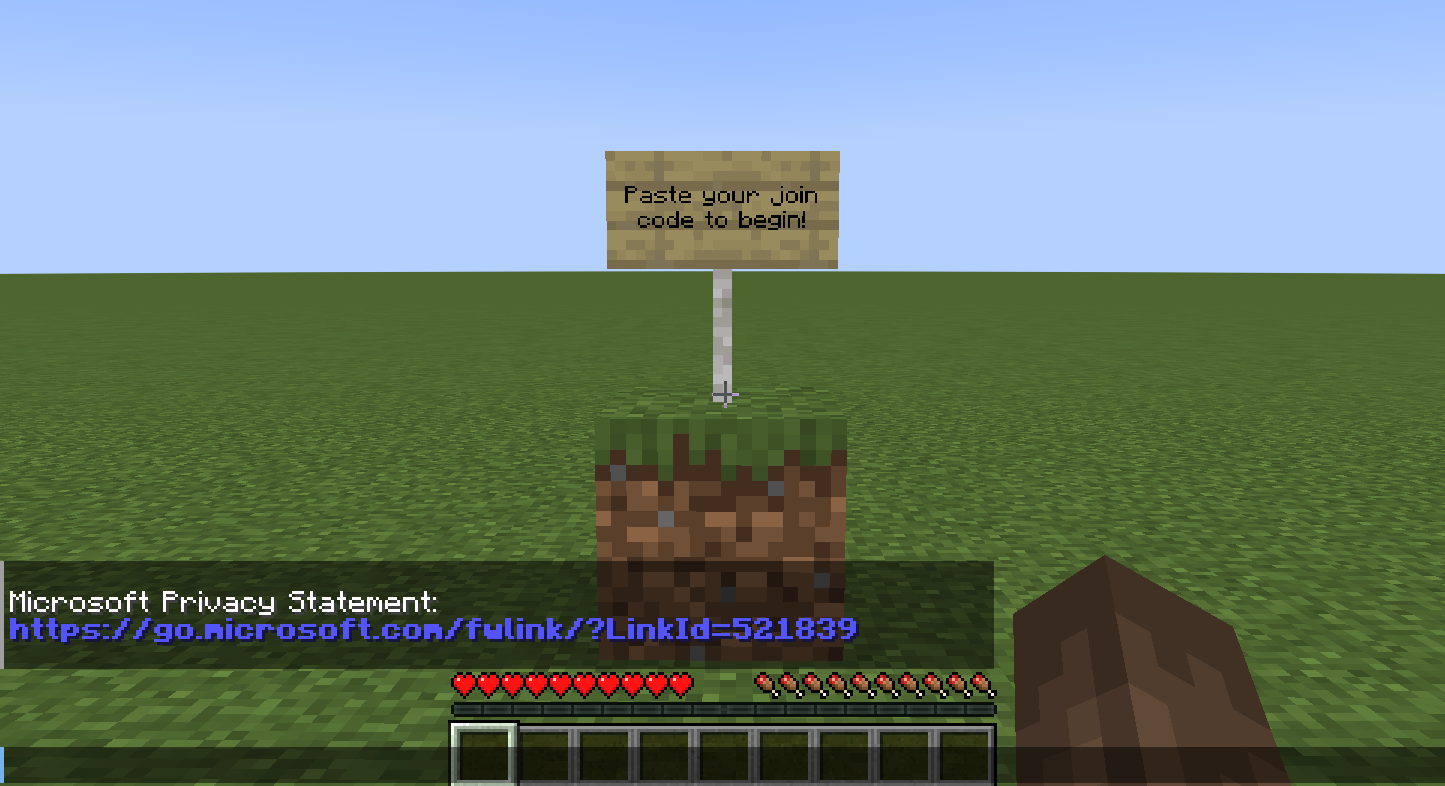}
    \caption{The human participant is \textit{spawned} in the Lobby world when they join the server. It's a flat world where the only action they're allowed to do is to paste a \textit{Join Code} in the chat box.}
\end{figure}

\begin{figure}[H]
    \centering
    \includegraphics[width=0.8\textwidth]{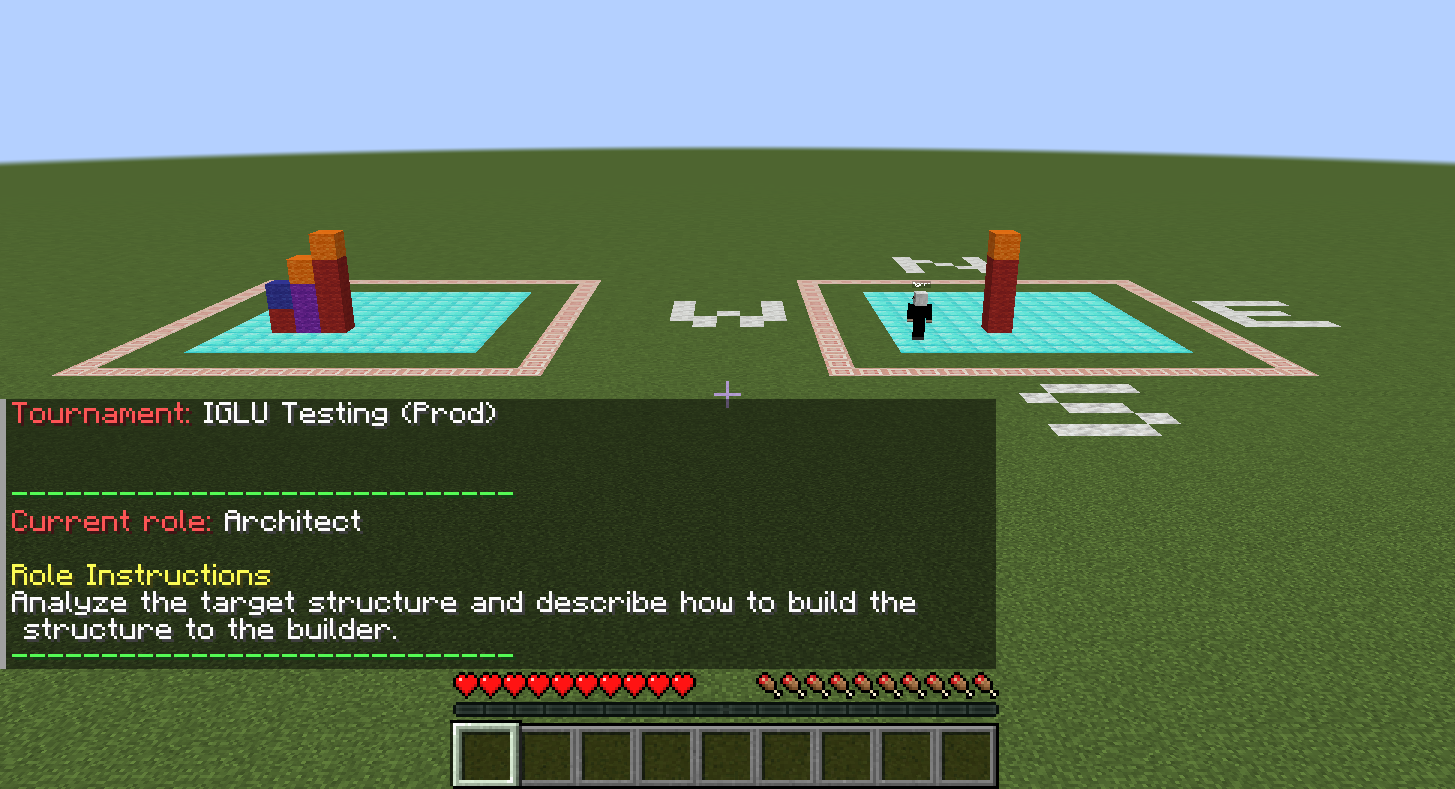}
    \caption{Initial view that the human participants see when they first join a game. The \textit{target structure} can be seen on the left side, and the agent and its initial structure can be seen on the right. The agent's \textit{build zone} has cardinal directions to make it easier for the human to provide instructions with absolute directions rather than having to rely on relative \textit{left, right}. At the start of the game, the participant is also provided with instructions detailing their role and goal for this session.}
    \label{fig:greenlands-new-game-start-screenshot-supplementary}
\end{figure}

\begin{figure}[H]
    \centering
    \includegraphics[width=0.8\textwidth]{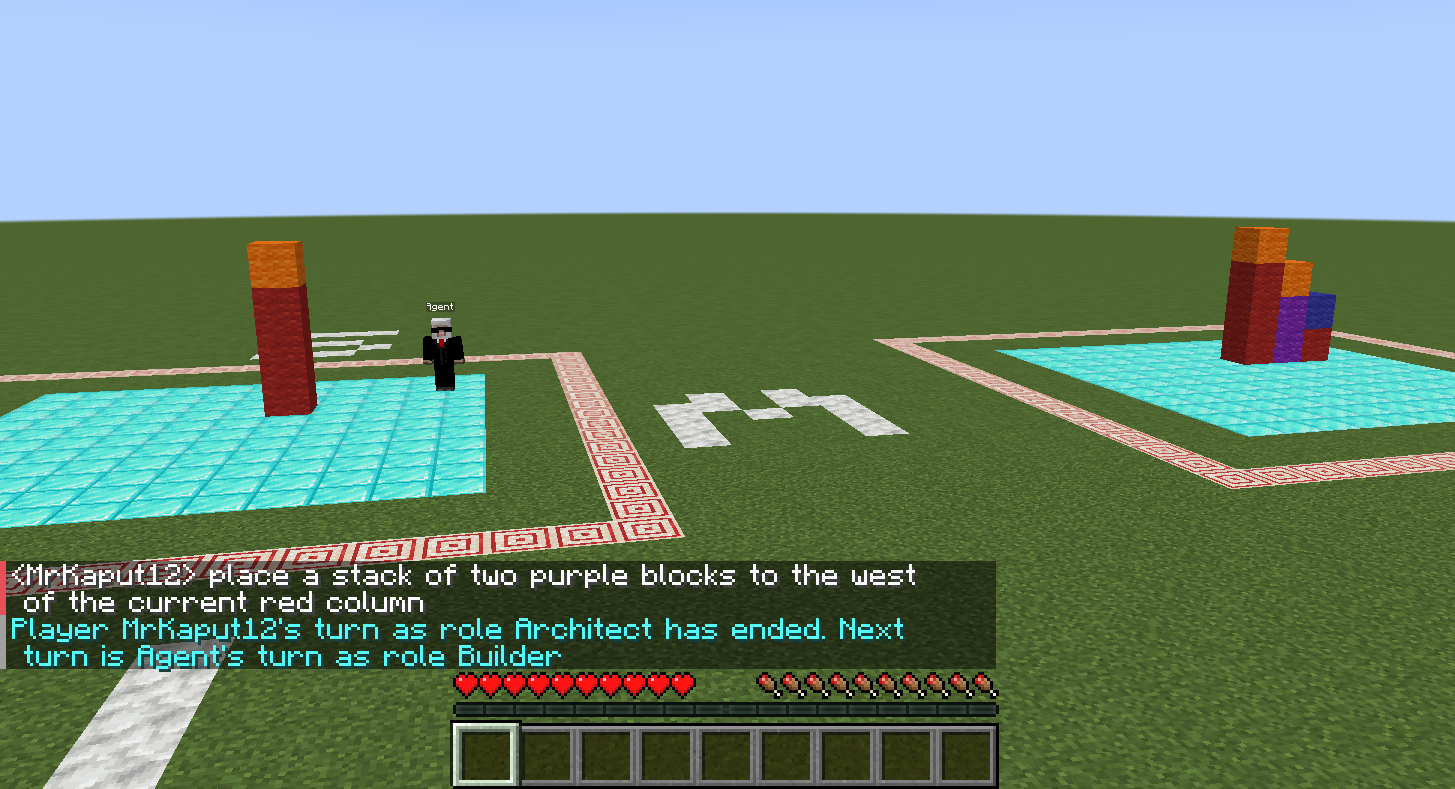}
    \caption{A human participant providing instructions to the agent (seen on the left side of the picture), and then ending their turn.}
\end{figure}

\begin{figure}[H]
    \centering
    \includegraphics[width=0.8\textwidth]{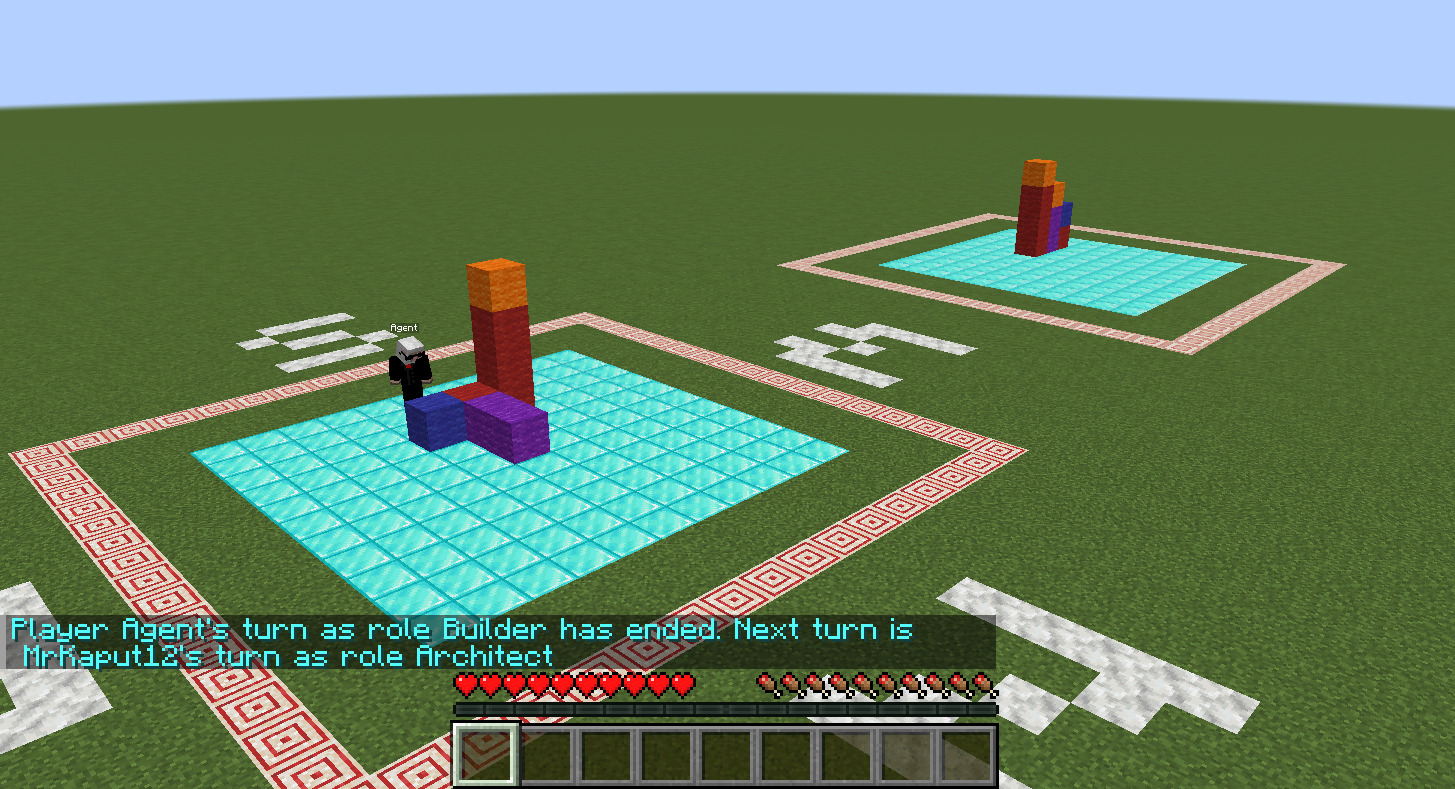}
    \caption{The agent has just performed its action in response to the human's instructions, and has now ended their turn.}
\end{figure}

\begin{figure}[H]
    \centering
    \includegraphics[width=0.8\textwidth]{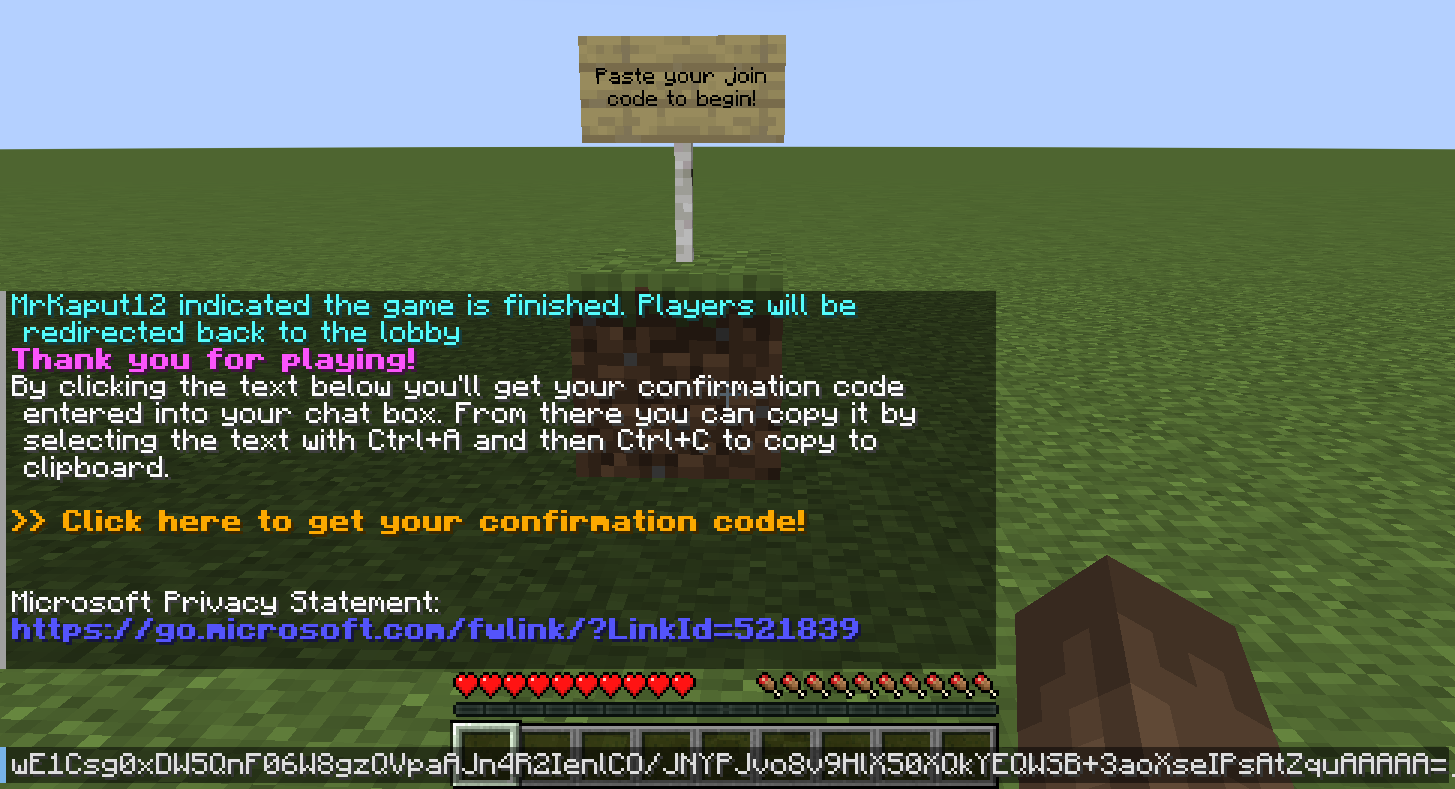}
    \caption{The human participant has finished the game and is sent back to the Lobby world. The chat box explains to them how to get the \textit{confirmation code} for the game.}
\end{figure}

\subsection{Examples of Human Annotations}
\label{sec:greenlands-human-annotations-examples-supplementary}

Below is a list of comments provided by the human participants for each of the agents. These have been manually chosen as representative of each agent's performance, and have been slightly paraphrased for correctness and readability. 

% breaking down comments per agent
\textbf{Brain Agent}
\begin{itemize}
    \item \textit{"The agent built a lot and was even able to break things but they were not able to choose colors right or destroy the right blocks."}
    \item \textit{"This agent was able to move around the structure and place blocks but it would always also instantly delete them. It was not able to figure out height also and kept building to high."}
    \item \textit{"The agent was largely unresponsive not really doing anything no matter the command whether it be to build or to break."}
    \item \textit{"It was able to build 3 blocks of blue like I wanted, but it was the wrong way. Then I wanted it to fix its mistakes, but the AI broke and started building and destroying blocks randomly."}
\end{itemize}

Here we can see that \textit{Brain Agent} tends to make actions even though it sometimes was unresponsive (ends its turn immediately without doing any action). It also tended to slightly obey the instructions of the user, especially during the first turn, but it would then start performing random actions.

\hfill

\textbf{MHB-Pegasus}
\begin{itemize}
    \item \textit{"The agent placed the blocks at the wrong location, ignores the location I ask them to place blocks at, kept building in the middle, and ignored my locations I was giving them."}
    \item \textit{"The agent was completely unresponsive not even really moving much just receiving commands and not acting on them at all."}
    \item \textit{"The agent placed a lot of blocks and got rather close to what was supposed to be the structure but they placed some wrong ones and could not destroy any blocks."}
    \item \textit{"The agent was able to place a lot of blocks but none that were part of my commands or even the right color at times. Along with that they did not even break them once placed."}
\end{itemize}

For the \textit{MHB-Pegasus} agent we again see it suffers from the \textit{unresponsiveness} problem. As with \textit{Brain Agent}, it seems to align to human instructions for the first few turns but later devolves into random action. 

\hfill

\textbf{MHB}
\begin{itemize}
    \item \textit{"The agent did not listen to my commands at all. It just did nothing. The first command had it back up and look down, then it refused to do anything else."}
    \item \textit{"The agent placed the blocks on the incorrect side of the grid. The agent followed part of my command, and placed 3 red blocks, but they were placed improperly at the wrong location."}
    \item \textit{"The agent was able to place and break blocks but did not follow any of my commands besides breaking the wrong blocks they placed."}
    \item \textit{"The agent was incapable of turning and was stuck building in one direction with the wrong colors."}
\end{itemize}

We can see that, overall, all three agents suffer from the same problems: ending up in a state where they can't decide on a next action and end their turn prematurely (even though the human clearly tells them what to do), obeying only part of the action (placing blocks in correct location but of different colors, or vice-versa), doing sensible actions only for the first few turns.

\end{document}